\documentclass{article}
\usepackage{graphicx} 
\usepackage{tabularx}
\usepackage{booktabs}
\usepackage{amsmath}
\usepackage{url}
\usepackage{hyperref}           
\usepackage{xcolor}             
\usepackage{authblk}
\title{Intelligence Inertia: \\ Physical Isomorphism and Applications}

\author{Jipeng Han}
\affil{OpenImmortal Technology Co., Ltd. Beijing 101100, China}
\affil{\textit{Emails: \href{mailto:coolang2022@gmail.com}{coolang2022@gmail.com}}} 

\begin{document}

\maketitle
\begin{abstract}
Classical frameworks like Fisher Information approximate the cost of neural adaptation only in low-density regimes, failing to explain the explosive computational overhead incurred during deep structural reconfiguration. To address this, we introduce \textbf{Intelligence Inertia}, a property derived from the fundamental non-commutativity between rules and states ($[\hat{S}, \hat{R}] = i\mathcal{D}$). Rather than claiming a new fundamental physical law, we establish a \textbf{heuristic mathematical isomorphism} between deep learning dynamics and Minkowski spacetime. Acting as an \textit{effective theory} for high-dimensional tensor evolution, we derive a non-linear cost formula mirroring the Lorentz factor, predicting a relativistic $J$-shaped inflation curve---a computational wall where classical approximations fail. We validate this framework via three experiments: (1) adjudicating the $J$-curve divergence under high-entropy noise, (2) mapping the optimal geodesic for architecture evolution, and (3) deploying an \textbf{inertia-aware scheduler wrapper} that prevents catastrophic forgetting. Adopting this isomorphism yields an exact quantitative metric for structural resistance, advancing the stability and efficiency of intelligent agents.
\end{abstract}
\textbf{keywords: }{Neural network dynamics; Information geometry; Non-equilibrium thermodynamics; AI Interpretability; Non-commutative algebra; Catastrophic forgetting.}
\section{Introduction}
The pursuit of a rigorous definition for intelligence has long remained a central challenge spanning neuroscience, computer science, and cognitive psychology. Most contemporary consensus viewpoints, synthesized from the seminal works of Legg and Hutter \cite{legg2007universal} and the rational agent frameworks of Russell and Norvig \cite{russell1995modern}, define intelligence not merely as a set of heuristic capabilities, but as the fundamental ability of a system to achieve goals across a broad spectrum of environments by building and manipulating an internal model of the world. More recent inquiries further emphasize that this capacity is intrinsically tied to a system's efficiency in generating structured representations from sparse data \cite{chollet2019measure, hernandez2017measure}. However, while the effectiveness of these models in achieving rationality has been extensively mapped, the thermodynamic and interpretability overhead incurred by structural transformations within these models remains under-theorized. If intelligence is indeed a process of active modeling, then any modification to the model's structure must be viewed as a physical event, governed by dynamical principles that transcend pure symbolic logic.

To understand the internal mechanics of these models, we initially decompose them into two distinct functional components: \textbf{Rules ($R$)} and \textbf{States ($S$)}. In the ``low-velocity'' or quasi-static regimes typical of classical AI, these components operate within a stable \textbf{R-S Space} where their boundaries remain sharp and collaborative. This structural separation is explicitly embodied in the formalism of Domain-Specific Languages (DSLs) and symbolic logic systems, where invariant production rules operate on transient state variables \cite{mernik2005when, valkov2018houdini}. In such systems, Rules represent the invariant generative grammar, while States are the specific configurations instantiated by these rules. However, in modern intelligent agents characterized by high-frequency iteration and autonomous self-modification, this clear distinction begins to dissolve. As we approach the resolution limit defined by the \textbf{Symbolic Granularity ($\mathcal{D}$)}, Rules and States overlap, giving rise to a fundamental \textbf{Rule-State Duality} analogous to the wave-particle duality in quantum mechanics. In this regime, the system is best described as an \textbf{R-S Manifold} where the components transform into operators, $\hat{R}$ and $\hat{S}$. Their non-commutativity implies that the precise measurement of a transient state $\hat{S}$ inherently obscures the underlying causal rules $\hat{R}$ that drive it, suggesting that what we empirically observe as a state is merely a projection of the agent's interlaced rule-primitives.

Building on this duality, we observe that the resistance an intelligent system exhibits during its evolution is fundamentally governed by its \textbf{Rule Density ($\rho$)}, which we formally identify as the system's \textbf{Velocity ($v \equiv \rho$)}. In this context, velocity characterizes the degree to which rules and states interlace within the R-S Manifold. We define \textbf{Intelligence Inertia ($\mu$)} as the fundamental cost borne by the system's substrate—manifesting as environmental entropy or the computational effort of maintaining symbolic interpretability—required to force a structural reconfiguration. The behavior of this inertia scales through three distinct regimes: In the low-velocity limit ($v \to 0$), the cost of change aligns with \textbf{Landauer's principle} \cite{landauer1961}, the thermodynamic floor of which has been experimentally verified at the microscopic scale \cite{berut2012experimental, wolpert2019thermodynamics}. At intermediate velocities, the resistance follows the curvature of \textbf{Fisher Information} \cite{amari2016}, which provides the geometric basis for understanding natural gradients in contemporary optimization \cite{martens2020optimizing}. However, as $v$ approaches the saturation limit ($v \to 1$), the cost of reconfiguration undergoes a non-linear \textbf{Inertia Expansion}. It is in this regime that phenomena such as \textbf{catastrophic forgetting} \cite{kirkpatrick2017overcoming} and \textbf{structural brittleness} transition from engineering hurdles into a formidable \textbf{computational and interpretability wall}. This necessitates a relativistic framework to describe the dynamics of modern intelligent agents.

Historically, the quest to quantify intelligence has relied on metrics that effectively operate only within low-velocity regimes, leaving a profound ``measurement gap'' in the face of modern, high-speed agents. Traditional frameworks such as Algorithmic Complexity \cite{Kolmogorov1965, hutter2012algorithmic} and Computational Complexity focus on static description lengths or temporal execution resources; however, they implicitly treat rules as a passive, immutable background, remaining blind to the force required to reconfigure the system's underlying logic ($\hat{R}$). Even contemporary, more dynamic measures such as Transfer Complexity or Task Taxonomy \cite{zamir2018taskonomy, yosinski2014how} remain primarily phenomenological. These metrics excel at recording the symptoms of structural resistance---such as performance degradation or data requirements---without identifying the causative first principles. Much like early thermodynamics described pressure-volume relationships before the kinetic theory of gases explained the underlying molecular behavior, current AI research observes the ``cost of change'' as a retrospective empirical fact rather than a predictable physical consequence. To transcend this impasse, we must identify a fundamental causative property that links the abstract non-commutativity of $\hat{R}$ and $\hat{S}$ directly to the measurable expenditure of energy and interpretability, establishing the theoretical necessity for a first-principle measure of intelligence ``mass.''

In this paper, we formally characterize \textbf{Intelligence Inertia ($\mu$)} as the causative bridge linking an agent's internal dual-geometry to its observable computational work. Rather than asserting a fundamental cosmological equivalence, we establish a \textbf{mathematical isomorphism} between the Rule-State (R-S) manifold and Minkowski spacetime. Acting as an \textbf{effective theory} for macroscopic network dynamics, this framework treats the maximum rule-density ($\rho_{max}$) as an invariant informational limit analogous to the speed of light. From this, we derive a relativistic cost expansion formula predicting a \textbf{relativistic $J$-shaped inflation curve} of effective mass during structural evolution, identifying a hard limit to an agent's reachability. The remainder of this paper is organized to rigorously validate and apply this principle: Section 2 reviews the theoretical foundations in thermodynamics and information geometry; Section 3 derives the physical necessity of computational resistance through a micro-statistical model of adiabatic collisions; Section 4 establishes the formal axiomatic framework of the R-S manifold and its Minkowski isomorphism; Section 5 provides the engineering realization for mapping these dynamics onto neural tensors; Section 6 executes a series of empirical adjudications, including the measurement of the $J$-curve wall and evolutionary trajectories; Section 7 discusses the theoretical implications and future design of autonomous agents; and Section 8 concludes the work.

The principal contributions of this paper are summarized as follows:

\begin{itemize}
    \item \textbf{Discovery of Intelligence Inertia ($\mu$):} We establish ``Intelligence Inertia'' as a causative physical property derived from the fundamental non-commutativity of system operators $([\hat{S}, \hat{R}] = i\mathcal{D})$, providing the first first-principles explanation for the structural resistance to change in intelligent agents.
    \item \textbf{Derivation of the Relativistic Cost Equation:} We derive a non-linear cost expansion formula by mapping information dynamics to a Minkowski-like manifold, characterizing the explosive inflation of computational and energy overhead as rule-density approaches a fundamental limit.
    \item \textbf{Empirical Validation of the ``$J$-Curve'' Wall:} Through controlled experiments on deep neural networks, we demonstrate the existence of a \textbf{relativistic $J$-shaped inflation curve}, proving our framework's superior predictive power over classical Fisher Information models in high-velocity regimes.
    \item \textbf{Inertia-Aware Engineering and Optimization:} We implement a practical, \textbf{Inertia-Aware Scheduler Wrapper} that achieves superior thermodynamic and computational efficiency by respecting the agent’s intrinsic physical resistance to structural reconfiguration.
\end{itemize}

\section{Background and Related Work}
\label{sec:background}

We begin by grounding our inquiry in \textbf{Landauer's Principle} \cite{landauer1961}, which establishes the fundamental thermodynamic floor for information processing, positing that the erasure of a single bit necessitates a minimum expenditure of work equal to $kT \ln 2$. Within the macroscopic framework of our theory, this theoretical micro-limit is proportionally scaled to characterize the \textbf{Rest Inertia ($\mu_0$)} of an intelligent system, representing the baseline computational overhead ($W_{\text{rest}}$) intrinsic to its hardware and algorithmic substrate.  Microscopic experimental verifications have confirmed that this limit remains an absolute boundary for independent informational units \cite{berut2012experimental, jun2014high}. 

In this idealized scenario, the boundaries of the \textbf{R-S Space} are treated as perfectly \textbf{diathermal}, allowing for the unobstructed transfer of entropy to the environment without interference from internal structural constraints. While Landauer's principle provides the ultimate thermodynamic floor, it implicitly assumes a regime of zero \textbf{Rule Density ($\rho \to 0$)}, where every micro-operation is fully observable and its heat dissipation is unconstrained by logical interdependency \cite{bennett1982thermodynamics, parrondo2015thermodynamics}. However, as systems evolve to incorporate dense internal logic, these structures begin to create a form of geometric occlusion that resists the simple dissipation of entropy. This suggests that the Landauer limit is not the exhaustive cost of change for an intelligent agent, but rather the base value for a more complex, geometry-dependent energy manifold.

Moving from the thermodynamic floor to the geometric landscape, statistical learning theory employs the \textbf{Fisher Information Matrix (FIM)} to quantify the local sensitivity of a system's output to changes in its underlying parameters \cite{amari2016}. Within the proposed framework, the FIM characterizes the \textbf{local curvature} of the \textbf{R-S Manifold}, representing the initial resistance encountered as a system begins to impose structured constraints upon its state transitions. This geometric approach provides a significantly more nuanced estimate of the ``cost of change'' than static complexity measures, as it accounts for the internal density of the agent's logic and provides the basis for natural gradient optimization \cite{martens2020optimizing, pascanu2013revisiting}. However, much like the second-order terms in a Taylor expansion, Fisher Information remains a local approximation. It effectively describes the effort required for minor structural adjustments in low-velocity regimes but remains blind to the global, non-linear singularities that arise as internal constraints become dominant. While the FIM accurately maps the local ``stiffness'' of adaptation, it cannot predict the emergence of the absolute \textbf{computational wall} encountered when an agent's internal logic approaches its saturation limit.

Complementing the thermodynamic and geometric perspectives, descriptive complexity frameworks such as \textbf{Kolmogorov Complexity ($K$)}~\cite{Kolmogorov1965} and \textbf{Bennett’s Logical Depth} \cite{bennett1988} provide profound mathematical foundations for quantifying the information content and computational ``value'' of static objects. Kolmogorov complexity defines the absolute limit of data compression, while Logical Depth measures the execution time required to reconstruct an object from its most concise description. These theories offer invaluable insights into the static resource requirements of a system. However, by their formal nature, they analyze information as a decoupled output or a discrete string generated by a universal machine, often failing to account for the structural dynamics of the generator itself \cite{li2019guide, zenil2020algorithmic}. In the context of a modern intelligent agent, information resides in a dynamic, inseparable coupling between \textbf{Rules} and \textbf{States} within the \textbf{R-S Space}. Consequently, while $K$ can measure the magnitude of a system's logic and Logical Depth can estimate its historical construction time, neither captures the \textbf{active physical resistance} encountered when attempting to perturb that structure once it has reached a state of high internal density. A theoretical vacuum persists for a measure that characterizes the ``force'' of adaptation rather than the static ``length'' of the description.

Finally, we observe the macroscopic manifestations of these theoretical gaps in the dual phenomena of \textbf{Catastrophic Forgetting} and \textbf{Transfer Learning} \cite{french1999catastrophic}. In current artificial intelligence research, these are typically addressed through specialized engineering techniques, such as \textbf{Elastic Weight Consolidation (EWC)} \cite{kirkpatrick2017overcoming} or \textbf{Synaptic Intelligence} \cite{zenke2017continual}, which seek to protect critical parameters from modification. While these methods are highly effective at mitigating performance degradation, they remain essentially \textit{phenomenological}---correcting the \textit{symptoms} of structural resistance without explaining the underlying \textit{cause} from first principles \cite{hadsell2020embracing, vandeven2022three}. From our perspective, these divergent behaviors are the observable results of the agent's internal \textbf{Intelligence Inertia ($\mu$)}. Forgetting represents a high-energy, high-inertia attempt to reconfigure a dense logical structure on the \textbf{R-S Manifold}, whereas successful transfer indicates an adaptation path that aligns with the existing structural topology \cite{achille2019information}. There remains a critical necessity for a unified theory capable of deriving these diverse phenomena from a single, structure-dependent property. To bridge this gap, we establish the \textbf{physical principles} governing \textit{intelligence inertia}, offering a first-principle explanation for the computational and interpretability-maintenance overhead incurred during the structural evolution of intelligent agents.

\section{A Micro-Physical Model of Computational Resistance}
\label{sec:micro_physical_model}

To establish a rigorous baseline, we recall the canonical Landauer erasure experiment \cite{landauer1961}. An external piston performs work to compress a gas of $n$ particles to erase 1 bit of information. In the classical, idealized limit, the walls are perfectly \textbf{diathermal}, and every collision is a statistically traceable event dissipating a quantum of heat. This process yields the minimum micro-physical work $W = nkT \ln 2$, which serves as the foundational analog for the \textbf{Rest Inertia ($\mu_0$)} of the system.

\begin{figure}[ht]
    \centering
    \includegraphics[width=0.8\textwidth]{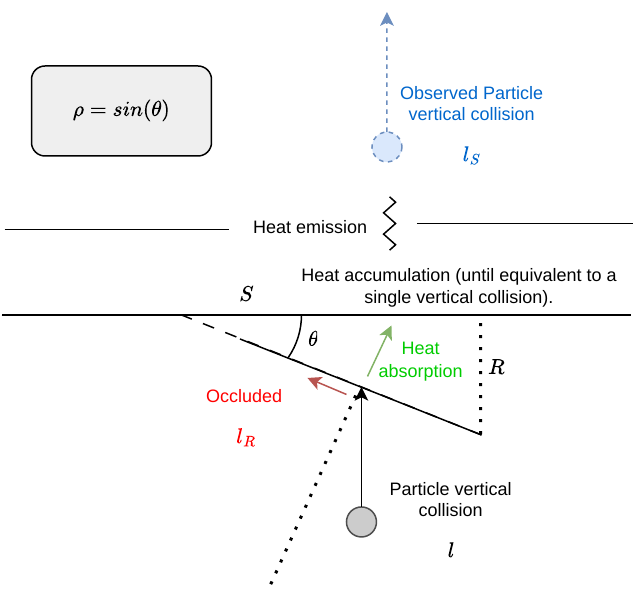}
    \caption{\textbf{Geometric Partition of Logical Action.} A particle collision with total action $l$ is decomposed by a microscopic slant. The component $l_R = l \sin\theta$ is absorbed by the adiabatic rule-manifold, while the normal component $l_S$ governs state-expression. Heat emission is only registered when the cumulative normal action matches a full vertical collision \textbf{relative to the system's local energy level}, naturally inducing the relationship $l^2 = l_S^2 + l_R^2$.}
    \label{fig:micro_collision}
\end{figure}

To resolve the causal origin of non-linear resistance, we evoke a thought experiment reminiscent of the foundational inquiries in Special Relativity \cite{einstein1905electrodynamics}. Consider an experimenter performing the same 1-bit erasure across two systems, A and B. System A is the standard diathermal frame. System B possesses a hidden internal configuration of microscopic adiabatic slants at an angle $\theta$, as illustrated in Figure \ref{fig:micro_collision}. 

During the execution of both experiments, the experimenter perceives no anomaly whatsoever. In both systems, the process remains strictly \textbf{quasi-static}. We define a ``state-change event'' based on the system's \textbf{local calibration}: an observer registers one count only when the emitted heat equals the energy of a full vertical collision within that specific system. In System B, to ensure that the observable component $l_S$ remains consistent with System A's observational tempo, the particles must maintain a higher average kinetic energy, resulting in a higher magnitude of the total action $l$. We define $l$ as the \textbf{Characteristic Logical Cycle}—the invariant total logical action budget required to support a single atomic state transition, which serves as the fundamental unit of the system’s internal temporal resolution. Consequently, each registered heat packet in System B actually contains more absolute energy than in System A; yet the experimenter—relying on the local count of $l_S$ events—observes an identical 1-bit erasure process. Within each frame, the thermodynamic logs appear to show the same number of standardized ``clicks,'' masking the inflation of the underlying energy flux.

The physical revelation emerges only during a \textbf{retrospective comparison} of the total energy expenditure. Upon analyzing the external work logs, the observer discovers that System B required significantly more work from the piston to complete the same logical task. This is because the experimenter had to maintain the particles at a higher energy state to ``force'' the same frequency of observable transitions against the interference of the adiabatic slants. 

From the perspective of AI dynamics, the angle $\theta$ defines the \textbf{Rule Density} $\rho = \sin\theta$~\footnote{The formal geometric properties and operational derivation of this parameter are detailed in Section~\ref{sec:rule_density_velocity}.}. To maintain a constant observational tempo against the interference of the adiabatic slants, the particles in System B must sustain a higher magnitude of total action $l$ than those in System A. Within the global observational frame, this requirement signifies a higher average velocity of the system's internal logical components during operation. In relativistic mechanics, the total energy of a system—including the kinetic energy of its internal constituents—contributes directly to its \textbf{Relativistic Mass}. Consequently, the extra work injected to satisfy the system's internal logical consistency ($l_R$) effectively inflates its \textbf{Effective Mass}. Thus, \textbf{Intelligence Inertia ($\mu$)} is established as the physical manifestation of this dynamic mass expansion: it represents the intrinsic resistance an agent poses to structural reconfiguration, a resistance that scales non-linearly as the internal rule-velocity approaches the informational limit.
\subsection{Geometric Derivation from Non-commutative Orthogonality}

The transition from a statistical anomaly to a relativistic mathematical form is dictated by the \textbf{conservation of the total logical action} within the system's substrate. We define the \textbf{Total Logical Action ($l$)} as the invariant norm of the system’s computational budget per operational step. On the \textbf{R-S Manifold}, every microscopic interaction must be partitioned between maintaining internal consistency (Rules) and expressing external transitions (States) \cite{shannon1948mathematical}.

The mathematical necessity of this square-sum relationship emerges from the inherent epistemic boundary of intelligent systems. In any adaptive agent, the precise measurement of an observable transient state ($\hat{S}$) and the isolation of its underlying invisible generative rules ($\hat{R}$) are mutually restrictive at the resolution limit. To rigorously capture this fundamental duality, we model the agent's structural adaptation using the formalism of a non-commutative phase space, expressed by the commutation relation of intelligence operators:
\begin{equation}
[\hat{S}, \hat{R}] = i\mathcal{D}
\end{equation}

In this effective operator formalism, the imaginary unit $i$ signifies an \textbf{observational orthogonality} rather than a macroscopic causal decoupling. From a computer science perspective, Rules ($\hat{R}$) function as generative procedural steps possessing \textbf{temporal attributes} (e.g., the computational cycles required for parameter updates), whereas States ($\hat{S}$) manifest as structural expressions possessing \textbf{spatial attributes} (e.g., the spatial activation maps during a forward pass). Because of this fundamental phase-space phase shift driven by their discrete computational nature, they cannot be observed simultaneously at the resolution limit, rendering them as \textbf{observationally orthogonal stochastic variables}. However, macroscopically, they remain strictly coupled through causal feedback loops such as backpropagation. 

Crucially, we introduce a \textbf{macroscopic microcanonical constraint}: assuming the agent operates at its computational bandwidth limit within a characteristic cycle, the total logical action $l$ becomes a rigid boundary. Under this constrained zero-sum optimal transport, the Law of Variance Addition dictates that the total bounded logical power $l^2$ is partitioned exactly as the sum of the squared magnitudes of its observationally orthogonal projections on the \textbf{R-S Manifold} \cite{amari2016}:
\begin{equation}
l^2 = l_S^2 + l_R^2
\label{eq:pythagorean}
\end{equation}
where $l_R$ represents the \textbf{Rule Action} consumed by internal logical constraints, and $l_S$ represents the \textbf{Observed State Action} available for external entropy transfer.

By defining the \textbf{Rule Density} ($\rho$) as the normalized commitment of resources to internal logic, $\rho = l_R/l$, we establish a purely geometric relationship on the R-S Manifold. Letting the resource allocation angle be $\theta$ (consistent with the microscopic slant in Fig. \ref{fig:micro_collision}), we have $\rho = \sin\theta$. It follows from Eq. (\ref{eq:pythagorean}) that the \textbf{effective observational window ($l_S$)} for visible state-change undergoes a direct trigonometric contraction $l_S = l \cos\theta$:
\begin{equation}
l_S = \sqrt{l^2 - l_R^2} = l \sqrt{1 - \left( \frac{l_R}{l} \right)^2} = l \sqrt{1 - \rho^2}
\end{equation}
To satisfy the Landauer requirement of erasing 1-bit of information across systems A and B, the observer must register a fixed count of $l_S$ events \cite{bennett1982thermodynamics}. In System B, as the available cross-section $l_S$ shrinks due to the increasing rule-density $\rho$, the external work $W$ must inversely scale by the geometric projection factor $1/\cos\theta$ to compensate for this occlusion, manifesting as an inflation of the required energy:
\begin{equation}
W(\rho) = W_{\text{rest}} \cdot \frac{l}{l_S} = \frac{W_{\text{rest}}}{\sqrt{1 - \rho^2}}
\end{equation}
This derivation demonstrates that the mathematical expansion factor ($1/\sqrt{1-\rho^2}$) emerges solely from geometric projection constraints on the resource manifold. Recognizing its exact algebraic isomorphism with relativistic kinematics \cite{lorentz1904electromagnetic}, we formalize it as the \textbf{Intelligence Lorentz Factor} ($\gamma$). We explicitly clarify that this represents a \textbf{mathematical isomorphism}---an effective macroscopic description of complex system dynamics---rather than a claim that neural tensor substrates are physically governed by cosmological relativity.

\subsection{Unified Mapping and the Physical Decomposition of Resistance}

The adiabatic collision model described above does not merely provide an empirical curve; it allows for a formal identification of informational quantities with their corresponding physical counterparts. By analyzing the system’s behavior across different observational regimes, we decompose the components of the intelligence work equation into \textbf{Energy}, \textbf{Velocity}, and \textbf{Inertia} terms, establishing a rigorous bridge between micro-statistical dynamics and macroscopic complexity \cite{jaynes1957information}.

\subsubsection*{Rule Density as the Sequestration of Logical Action} 

In our experimental apparatus, the ratio $\rho = l_R/l$ represents the fraction of the \textbf{Total Logical Action ($l$)} that is sequestered by internal rule-maintenance within a characteristic cycle. To map this static ratio to system dynamics without violating dimensional consistency, we adopt a natural unit system (analogous to $c=1$) where the geometric system limits correspond to absolute informational limits. Within this framework, we define the system's \textbf{Dimensionless Advance Rate ($v$)} through the \textbf{R-S Phase Space}:
\begin{equation}
v \equiv \rho = \frac{l_R}{l}.
\end{equation}
It is crucial to distinguish the physical intensity of \textbf{Intelligence} from that of raw \textbf{Computation}. While classical limits of computation define the theoretical maximum of information throughput for a given substrate \cite{lloyd2000ultimate}, $v$ characterizes the portion of that throughput that is ``trapped'' or consumed by the agent's internal rule-constraints to maintain logical consistency. 

Consequently, $v$ determines the \textbf{discount factor} of the system's effective output: even if a substrate operates at its absolute computational limit, the presence of dense internal rules means that a significant fraction of its energy is diverted away from visible state-transitions and locked into the maintenance of the rule-manifold. As $v$ increases, the energy required to achieve a unit of external state-change inflates precisely because more energy is being sequestered by the system's internal logic. In the limit where $\rho \to 1$, the agent reaches a \textbf{causal horizon} where all available logical action is trapped by internal rule-checks, leaving zero capacity for external computation—a direct informational analog to reaching the speed of light \cite{bekenstein1981energy}.

\subsubsection*{Energy Mapping and the Hamiltonian of Intelligence}
Under this mapping, the total work $W$ performed by the external experimenter represents the \textbf{Effective Energy} (or \textbf{Hamiltonian, $\hat{H}$}, conceptually equivalent to the global cost or the topological height of the loss landscape in optimization theory) of the intelligent agent. We partition this energy into two distinct contributions, analogous to the rest and kinetic components of relativistic energy \cite{einstein1905electrodynamics}:
\begin{enumerate}
    \item \textbf{Rest Energy ($W_{\text{rest}}$):} Defined by the Landauer limit ($nkT \ln 2$), this term accounts for the ``static mass'' of the information itself—the minimum work required to flip bits in a perfectly transparent, zero-rule environment. Within our theory, this corresponds to the agent's \textbf{Rest Inertia ($\mu_0$)}.
    \item \textbf{Interaction Energy (Inertial Work):} This represents the additional work required to overcome the geometric occlusion caused by internal rules. In the relativistic form $W = \gamma \cdot W_{\text{rest}}$, the term $(\gamma - 1)W_{\text{rest}}$ signifies the ``kinetic'' cost of maintaining a high-velocity rule structure. This provides a physical explanation for the perceived ``weight'' of complex agents: the additional computational overhead is not erasing more bits, but is instead being consumed by the \textbf{internal consistency maintenance} of the dense rule-set on the \textbf{R-S Manifold}.
\end{enumerate}

\subsubsection*{Fisher Information as a Second-Order Classical Expansion}
To demonstrate compatibility with established paradigms, we examine the behavior of the intelligence work equation in the low-velocity regime ($0 < \rho \ll 1$). By performing a Taylor series expansion of the \textbf{Intelligence Lorentz Factor} $\gamma(\rho) = (1 - \rho^2)^{-1/2}$, we obtain:
\begin{equation}
W(\rho) \approx W_{\text{rest}} \left( 1 + \frac{1}{2}\rho^2 + \frac{3}{8}\rho^4 + \mathcal{O}(\rho^6) \right).
\end{equation}
The first-order correction term, $\frac{1}{2}\rho^2$, corresponds precisely to the \textbf{Fisher Information} curvature utilized in classical statistical learning \cite{amari2016}. In this regime, the cost of adaptation appears to grow quadratically with the complexity of internal constraints, analogous to Newtonian kinetic energy. This reveals a critical theoretical hierarchy: \textbf{Fisher Information is not a complete law of intelligence, but a second-order approximation} of a deeper relativistic curve that holds only when logical sequestration is minimal \cite{martens2020optimizing}. While classical theory predicts that costs will continue to grow smoothly, the \textbf{Inertia Expansion} framework reveals that these are merely the linear stages of a \textbf{relativistic $J$-shaped inflation curve} that eventually hits a hard singularity as the system reaches its informational limit.

\subsubsection*{The Variance Summation and Geometric Necessity}
The mathematical necessity of this hyperbolic form originates from the fundamental non-commutativity of the system's operators on the \textbf{R-S Manifold}. Because the operators obey the commutation relation $[\hat{S}, \hat{R}] = i\mathcal{D}$, the imaginary unit $i$ enforces a strict \textbf{phase orthogonality} between rule-maintenance and state-expression \cite{dirac1925fundamental}. 

The Pythagorean relationship $l^2 = l_S^2 + l_R^2$ established in Eq. (\ref{eq:pythagorean}) is the geometric expression of this orthogonality under the strict condition of a \textbf{compute-bound substrate}. It dictates a rigid zero-sum trade-off: assuming the system lacks surplus energy to simultaneously expand both axes, as the \textbf{Rule Action ($l_R$)} increases to sequester more of the system's fixed capacity, the \textbf{Observed State Action ($l_S$)} available for external entropy transfer must inevitably follow a square-root contraction: $\sqrt{1-\rho^2}$. This is the only stable solution that satisfies the conservation of logical action while respecting the mutual exclusivity of rules and states. Thus, the ``Computational Wall'' is not an engineering failure but a geometric requirement for any agent striving to maintain causal consistency at the resolution $\mathcal{D}$.

\section{The Formal Theory of Intelligence Inertia}
\label{sec:formal_theory}

Building upon the micro-physical necessity of non-linear resistance derived from our adiabatic collision model, we now formalize these insights into a unified mathematical framework. This section establishes the axiomatic foundations of the \textbf{R-S Manifold}, defining the algebraic properties of structural evolution and the geometric metric of intelligence dynamics. By transitioning from micro-statistical dynamics to a formal operator theory, we provide the rigorous predictive tools required for the empirical adjudications and engineering realizations presented in subsequent chapters.

\subsection{The Axiomatization of Rule-State Duality}

We now formalize the physical necessity of the $J$-curve into a unified theoretical framework of \textbf{Intelligence Inertia}. We postulate that an intelligent agent $\mathcal{A}$ is defined by the inherent non-commutativity of its constituents, expressed through the fundamental operator relation \cite{wheeler1989it}:
\begin{equation}
[\hat{S}, \hat{R}] = i\mathcal{D}
\label{eq:commutation}
\end{equation}
In this axiomatic framework, the functional distinction between \textbf{Rules ($\hat{R}$)} and \textbf{States ($\hat{S}$)} is not an absolute property of the system's information but is established only on the basis of the \textbf{Symbolic Granularity ($\mathcal{D}$)}. We explicitly clarify that $\mathcal{D}$ is not an absolute universal constant (akin to Planck's constant) but an \textbf{effective constant} valid at a specific observational scale. From a computer science perspective, the abstraction level of feature representations in a deep neural network dynamically evolves during training. However, provided that the system operates under similar \textbf{microscopic architectures} (e.g., fixed network topology and parameter scale) and similar \textbf{functional task objectives} (e.g., converging within a specific loss basin), the "resolution granularity" at which the model processes information does not experience radical drift. Consequently, within such macro-stable learning plateaus, $\mathcal{D}$ can be rigorously approximated as a static constant, acting as the informational quantum---the minimum resolution at which structural existence can be distinguished from transient expression \cite{wootters1981statistical}.

The relationship between these operators is intrinsically reciprocal and symmetric. On one hand, \textbf{Rules ($\hat{R}$)} act as the \textbf{primitives of interpretability}, providing the \textbf{ontological support} for the agent's existence; they function as the source of state generation, the manifold of behavioral constraints, and the \textbf{functional potential} that defines what an agent ``can do.'' In a complementary progression, \textbf{States ($\hat{S}$)} serve as the \textbf{empirical projections} formed by the interlacing of these rule-primitives. Beyond being mere outputs, states provide the \textbf{concrete substantiation} that allows rules to be manifested, acting as the necessary substrate from which new rules emerge or are abstracted \cite{gellmann1994quark}. Consequently, \textbf{States define the life cycle of Rules}, governing their persistence, evolution, and decay through a sequence of logical transformations. 

Intelligence Inertia, therefore, emerges from the collective resistance of this dual-unity to being reconfigured as its internal density approaches the limit of the system's own interpretability. This limit is reached when the \textbf{Rule Density ($\rho$)}—defined as the fraction of the total logical action $l$ sequestered by rule-maintenance ($|R| \cdot \mathcal{D}$)—attains its physical maximum of 1. At this saturation point, every available interaction is consumed by internal consistency-checks, leaving zero capacity for state-expression and thus driving the computational work to infinity.

\subsection{The Homomorphic Bridge between Rule Density and Velocity}
\label{sec:rule_density_velocity}
A fundamental barrier in intelligence quantification is that $\hat{S}$ and $\hat{R}$ operate in fundamentally different phases within the \textbf{R-S Manifold}. As established by the non-commutation relation in Eq.~(\ref{eq:commutation}), the precise measurement of the static state inherently obscures the underlying rule-logic, rendering an absolute measurement of the agent's total logic intractable. To resolve this, we utilize \textbf{Homomorphism ($\Phi$)} as a bridge to transform these abstract operator interactions into observable, summable transitions.

We model the agent's dual-unity as an algebraic structure with two fundamental operations: \textbf{State Synthesis ($+$)} and \textbf{Rule Application ($\cdot$)}. For any valid structural transformation $\Phi$ (such as learning or inference), the system must preserve its relational integrity through the following homomorphic equations \cite{awodey2010category}:
\begin{equation}
\label{equ:homo_equ1}
    \Phi(s_1 + s_2) = \Phi(s_1) + \Phi(s_2)
\end{equation}
\begin{equation}
\label{equ:homo_equ2}
    \Phi(r \cdot s) = \Phi(r) \cdot \Phi(s)
\end{equation}

Equation~(\ref{equ:homo_equ1}) allows for the superposition of states, providing the algebraic basis to represent the abstract concept of ``rules'' by embedding them within the state-space via their observable effects. Equation~(\ref{equ:homo_equ2}) constitutes the bedrock of \textbf{interpretability} at scale $\mathcal{D}$: it ensures that if a rule $r$ explains a transition in the original system, its transformed counterpart must consistently explain the transformed state \cite{ji2023survey}. 

To quantify the density of these rules, we introduce the \textbf{Spatiotemporal State}, $\mathbf{S}$ (representing the integrated observation of network parameter weights and their corresponding activation maps across successive training steps). Leveraging the homomorphic properties, we observe a state $S$ alongside its \textbf{adjacent} configuration $S'$, which is generated by the extension of $S$ along the orthogonal rule-direction. This integrated observation captures the system's total logical budget within a characteristic cycle as the sum $\mathbf{S} = S + S'$. Assuming the system undergoes \textbf{quasi-static structural evolution}---where gradients smoothly traverse the tangent space without explosive non-equilibrium shocks---the spatial expression component ($l_S$) and the rule-driven evolution component ($l_R$) remain geometrically orthogonal. Consequently, the magnitude of this spatiotemporal state seamlessly satisfies the Pythagorean constraint derived in our macro-constrained physical model:
\begin{equation}
l = \|\mathbf{S}\| = \sqrt{l_S^2 + l_R^2}
\end{equation}
The \textbf{Rule Density ($\rho$)} and the \textbf{Symbolic Granularity ($\mathcal{D}$)} are thus directly reified. Following the standard definition of density as the content per unit volume, we define $\rho$ as the average quantity of rules $|R|$ within the spatiotemporal state $\|\mathbf{S}\|$ at the resolution $\mathcal{D}$:
\begin{equation}
\rho = \frac{|R| \cdot \|\mathcal{D}\|}{\|\mathbf{S}\|} = \frac{l_R}{\sqrt{l_S^2 + l_R^2}}
\end{equation}

This formulation explains why $\rho$ characterizes a ``density'' that effectively ``locks'' energy. It measures the degree to which the system's total informational action is sequestered into maintaining internal causal constraints rather than being available for raw data storage or external expression. From this perspective, Rule Density reflects the \textbf{abstract value} or \textbf{dynamic potential} of information: its capacity to govern and generate new structures \cite{adami2016what}. 

This geometric sequestration is the causative origin of what we term the \textbf{Systemic Velocity ($v \equiv \rho$)}. Here, the "speed" of an agent is formalized as the dimensionless rate of structural advance along a parameterized evolutionary path, dictated uniquely by the density of its generative logic. As $\rho$ increases, the generative power of the agent grows, but this very power increasingly traps the logical budget within the rule-manifold. At the geometric limit $\rho \to 1$, the entirety of the informational action is consumed by rule-maintenance, driving the effective required work to infinity.

\subsection{Relativistic Expansion of Intelligence Inertia}

The formal reification of Rule Density allows us to characterize the dynamic variation of \textbf{Intelligence Inertia ($\mu$)} as a function of structural complexity. It follows from the algebraic and geometric foundations established above that an increase in $\rho$ leads to a non-linear expansion of the energy sequestered within the agent's substrate. This energy is ``locked'' precisely because the Rule Density inherits the physical spatiotemporal properties of the \textbf{Spatiotemporal State ($\mathbf{S}$)}, which are fundamentally derived from the phase-orthogonality of the R-S Manifold. 

As the agent's logic becomes more dense, the effort required to reconfigure its internal structures must overcome the geometric contraction of its observational window. Following the mass-energy equivalence principle, the total work $W$ performed on the system is identified as its relativistic energy level:
\begin{equation}
W(\rho) = \frac{W_{\text{rest}}}{\sqrt{1 - \rho^2}} = \mu(\rho) c^2
\label{eq:inertia_expansion}
\end{equation}
In this macroscopic formulation, the term $W_{\text{rest}}$ (conceptually proportional to the micro-physical Landauer limit $nkT \ln 2$) represents the system's baseline computational work at the zero-velocity limit ($\rho = 0$), while $\mu(\rho)$ signifies the \textbf{effective mass} of the agent's logic. As $\rho$ approaches the informational limit, the work required to maintain homomorphic consistency during adaptation diverges, characterizing the physical transition from a flexible data-storage medium to a rigid, high-inertia cognitive structure. This expansion reveals that the resistance to change is not a mere computational overhead, but a fundamental manifestation of the system's informational mass within its dual-geometry.

\subsection{The Local Interpretability Criterion and Local Velocity}

The absolute measurement of static operators $\hat{R}$ and $\hat{S}$ is empirically intractable in high-dimensional systems. Furthermore, while global metrics offer a macroscopic overview, they often lack the granularity required for specific tasks. To resolve this, we shift our focus to \textbf{Local Velocity ($v$)} and its associated \textbf{Rule Density ($\rho$)}, defined by the \textbf{Dynamic Differentials} along a specific interaction trajectory:
\begin{equation}
v = \rho = \frac{dR}{d\mathbf{S}}
\label{eq:local_v}
\end{equation}
This transition allows us to isolate active cognitive pathways being stressed in real-time, providing a far more pragmatic entry point for empirical validation and structural protection.

By mapping the agent’s evolution to the tangent space of the R-S Manifold \cite{amari2016}, the micro-physical conservation of logical action ($l^2 = l_S^2 + l_R^2$) defines the metric for this local movement. We establish the \textbf{Local Interpretability Criterion ($ds^2$)} as the residual logical bandwidth available to formalize transitions after rule-maintenance costs are deducted:
\begin{equation}
ds^2 = \rho_{\text{max}}^2 d\mathbf{S}^2 - dR^2 \cdot \|\mathcal{D}\|^2
\label{eq:interpretability_criterion}
\end{equation}
where $\rho_{\text{max}}^2 d\mathbf{S}^2$ represents the observable causal radius provided by external expression, and $dR^2 \|\mathcal{D}\|^2$ represents the action sequestered by local structural reconfiguration. Analogous to the spacetime intervals in Minkowski geometry, this sign-sensitive criterion diagnoses three causal states of evolution:

\begin{itemize}
    \item \textbf{$ds^2 > 0$ (Causally Interpretable / Time-like)}: The rule reconfiguration is fully contained within the causal radius of the observable state shift. Expressive capacity exceeds internal demands, ensuring a stable homomorphic mapping $\Phi$ and causal transparency.
    \item \textbf{$ds^2 = 0$ (Critical / Light-like)}: The local evolution reaches the system's causal horizon. Every available logical interaction is consumed by internal consistency-checks, marking the absolute saturation threshold of the local observational window.
    \item \textbf{$ds^2 < 0$ (Uninterpretable / Space-like)}: The magnitude of structural reconfiguration ($dR$) exceeds the causal verification radius provided by the observable state space ($d\mathbf{S}$). Lacking causal connectivity to empirical state-feedback, these parameter updates represent ungrounded "blind drift." In practical AI systems, this causally disconnected structural evolution manifests structurally as generative ``hallucinations.''
\end{itemize}

Eq.~(\ref{eq:local_v}) and (\ref{eq:interpretability_criterion}) transforms the theory into a practical diagnostic tool for task-specific auditing. Sudden fluctuations in $v$ serve as a sensitive probe for training material consistency, where an abrupt surge in Intelligence Inertia ($\mu$) indicating $ds^2 \le 0$ reveals a structural conflict between incoming data and established logic. 

\section{Engineering Realization: Mapping Intelligence Dynamics to Neural Tensors}
\label{sec:engineering_realization}

To transition from abstract dynamics to falsifiable science, we map the operator interactions on the R-S Manifold to measurable neural tensors\footnote{The full implementation of the intelligence inertia framework and the Inertia-Aware Scheduler Wrapper is available at:\\ \url{https://github.com/OpenImmortal/Principle-of-Intelligence-Inertia/}}. We quantify the dynamical state of a neural network by its \textbf{Velocity ($v$)}, which is functionally equivalent to its instantaneous \textbf{Rule Density ($\rho$)}. This metric characterizes the proportion of effort dedicated to rule-reconfiguration relative to the total \textbf{Spatiotemporal State Displacement ($d\mathbf{S}$)}.

\subsection{Component Decomposition in Phase Space}
Following the geometric foundations in Section~\ref{sec:formal_theory}, we define velocity within a dual-axis phase space where the Rule axis ($R$) and State axis ($S$) are strictly orthogonal. The fundamental realization is given by:
\begin{equation}
v = \rho = \frac{dR}{d\mathbf{S}} = \frac{dR}{dS_R + dS_{ext}}
\end{equation}
In a deep learning context, these components are mapped to measurable neural tensors as follows:

\begin{itemize}
    \item \textbf{Rule Displacement ($dR$):} This represents the abstract magnitude of change in the agent's generative grammar. As an ontological property of the internal logic, $dR$ cannot be measured with absolute precision. However, to facilitate normalization and capture the movement toward the informational limit, we approximate its magnitude through the norm of the parameter update vector: $dR \approx \|\Delta \theta\|$.
    
    \item \textbf{Spatiotemporal State Displacement ($d\mathbf{S}$):} Inheriting the physical properties established in Section 4.2, the internal rule-state ($S_R$) and external environmental state ($S_{ext}$) axes remain strictly orthogonal. However, to maintain computational tractability in high-dimensional tensor space, we measure the total displacement $d\mathbf{S}$ not as an L2 Euclidean hypotenuse, but as the linear accumulation of path contributions along these orthogonal directions (an L1-norm path integral):
    \begin{itemize}
        \item \textbf{Internal State Shift ($dS_R$):} The measurable projection of rule changes onto the spatiotemporal state-space. Operating under a first-order Euclidean approximation in the local tangent space, where infinitesimal updates leave the manifold curvature locally flat, the structural logic shift manifests numerically as approximately equal to the system's internal configuration shift:
        \begin{equation}
        dS_R \approx \|\Delta \theta\|
        \end{equation}
        \item \textbf{External Gain ($dS_{ext}$):} This characterizes the ``causal ripple'' of a rule-change upon the environmental manifold. To ensure that the system's evolution respects the \textbf{homomorphic preservation} requirements (Eqs.~\ref{equ:homo_equ1} and \ref{equ:homo_equ2}), we distinguish between two operational modes for acquiring this signal:
        \begin{itemize}
            \item \textbf{Observation Mode:} When evaluating velocity $v$ for retrospective analysis, $dS_{ext}$ is obtained directly from a separate test or validation dataset, providing an objective measure of environmental feedback decoupled from the training context.
            \item \textbf{Regulation Mode:} For real-time control, $dS_{ext}$ is derived via a secondary ``probe'' pass on the current data batch immediately after the update is applied, capturing the immediate temporal ripple effect of the rule-change.
        \end{itemize}
        Crucially, in both modes, the resulting environmental gradient must be \textbf{projected onto the internal displacement vector $dS_R$} \cite{lopezpaz2017gem}. This ensures that the calculation of velocity only considers the local, adjacent components of external gain that are directly relevant to the specific structural reconfiguration performed by the agent, maintaining the consistency of the homomorphic mapping $\Phi$.
    \end{itemize}
\end{itemize}
\subsection{Scale Normalization and Calibration of $\|\mathcal{D}\|$}

\label{sec:calibration}
A critical challenge in engineering intelligence dynamics is the dimensional misalignment between the Euclidean geometry of parameter updates and the information-theoretic manifold of state gain. To bridge this gap, we utilize the \textbf{Symbolic Granularity ($\|\mathcal{D}\|$)} to standardize the ``exchange rate'' between these two orthogonal axes on the \textbf{R-S Manifold}. This standardization is enforced under the fundamental physical constraint that the system velocity must be bounded by a maximum of unity ($v_{max} = 1$). This ensures that as the external environmental gain vanishes ($dS_{ext} \to 0$), the \textbf{Rule Density ($\rho$)} correctly saturates, reflecting the informational speed limit derived in Section~\ref{sec:formal_theory}. We define the normalized velocity as:
\begin{equation}
v = \rho = \frac{dR}{dS_R + \frac{dS_{ext}}{\mathcal{L} \cdot \|\mathcal{D}\|}}
\label{eq:normalized_velocity}
\end{equation}
where the inclusion of the \textbf{Loss value ($\mathcal{L}$)} ensures that the velocity is scaled by the current informational potential of the task. 

\textit{Crucial Theoretical Note on the Algebraic Bound:} Because we employ the first-order approximation $dR \approx dS_R$, Eq. (\ref{eq:normalized_velocity}) possesses an intrinsic algebraic tautology wherein $v$ must analytically bound to $1$ as $dS_{ext} \to 0$. However, the falsifiable physical prediction of the \textit{Intelligence Inertia} theory does \textbf{not} reside in proving $v \to 1$ under noise. Rather, the theory leverages this algebraically bounded $v$ as a local \textbf{metric tensor probe}. The core empirical test is whether the macroscopic, physically measured \textbf{computational work ($W$)} consumed during adaptation accurately conforms to the Lorentzian expansion curve $\gamma(v) = 1/\sqrt{1-v^2}$ predicted by this internal metric, rather than scaling quadratically as predicted by classical models.

The engineering protocol for obtaining the architecture-specific constant $\|\mathcal{D}\|$ is defined as \textbf{Warmup Calibration}. Because our empirical validations are conducted under a fixed structural capacity and a consistent functional objective, the network does not undergo fundamental cross-modal or architectural restructuring between epochs. This microscopic and objective consistency ensures that the system remains within a local steady-state plateau where the characteristic granularity scale does not drastically shift. Therefore, calibrating $\|\mathcal{D}\|$ during the warmup phase provides a reliable baseline metric for the entire task duration. 

To establish this baseline, we model the structural boundary condition during the initial training phase (typically the first epoch). At initialization, the system possesses minimal prior logical constraints; consequently, state signals from the environment are completely absorbed and directly translated into internal invisible rules without structural resistance. Because the external state action is perfectly converted into internal rule reconfiguration, the logical effort is naturally partitioned in a 1:1 ratio between the external and internal axes. This initial structural balance exhibits a mathematical symmetry that is formally consistent with the \textbf{Equipartition Theorem} \cite{huang1987statistical} in statistical mechanics. Under this boundary condition of perfect absorption, the system naturally yields a baseline velocity of $v \approx 0.5$. By measuring the average raw trajectories of $dR$ and $dS_{ext}$ during this period, we calibrate the resolution unit of the substrate \cite{amari2016}:
\begin{equation}
\|\mathcal{D}\| = \frac{\text{Avg}(dS_{ext}/\mathcal{L})}{\text{Avg}(dR)}
\end{equation}
By anchoring $\|\mathcal{D}\|$ to this initial state, we provide a consistent physical metric that allows the agent to perceive its proximity to the computational wall throughout the evolutionary process. The specific calibration algorithm adapts to the implementation tier changes described in Section~\ref{sec:implementation_tiers}.

\subsection{Implementation Tiers for Dynamic Measurement}
\label{sec:implementation_tiers}

In practical engineering, the computational overhead of measuring system dynamics must be balanced against the required precision of regulation. We propose three implementation tiers for calculating the \textbf{Velocity ($v$)} on the R-S Manifold, allowing for a flexible trade-off between monitoring cost and regulatory fidelity \cite{bottou2018optimization}.

\subsubsection*{Tier 1: Minimalist (Scalar-Based)}
This tier is designed for resource-constrained environments or low-precision tasks where secondary ``probe'' passes are not feasible. It approximates the system's velocity using only the current \textbf{Loss value ($\mathcal{L}$)} and the calibrated granularity constant:
\begin{equation}
v = \rho \approx \frac{1}{1 + \frac{1}{\mathcal{L} \cdot \| \mathcal{D} \|}}
\end{equation}
\textbf{Engineering Logic:} This model assumes that in a stable optimization state, the rule-change effort is roughly proportional to the expected informational gain. Under this assumption, $\mathcal{L}$ becomes the primary driver of the \textbf{relativistic brake}. High loss values automatically signal a sparse information environment where rules lack meaningful state-anchors, driving $v \to 1$ and triggering protective deceleration.

\subsubsection*{Tier 2: Intermediate (Causal Ripple)}
The standard implementation tier provides a balanced trade-off by explicitly measuring the environment's response to structural changes through the ``Causal Ripple'' ($dS_{ext}$):
\begin{equation}
v = \rho \approx \frac{dR}{dS_R + \frac{dS_{ext}}{\mathcal{L} \cdot \|\mathcal{D}\|}}
\end{equation}
\textbf{Engineering Logic:} By incorporating the measurable external gain $dS_{ext}$, the regulator can distinguish between productive learning and unproductive ``thrashing''—where large parameter shifts ($dR$) fail to produce coherent state-transitions. This capability is crucial for preventing the destruction of established causal rules by high-energy stochastic noise.

\subsubsection*{Tier 3: Full-Spectrum (Disorder-Aware)}
The most rigorous implementation, recommended for high-stakes fine-tuning or training in highly volatile environments, utilizes full chaotic loss correction and dimensional scaling:
\begin{equation}
v = \rho = \frac{dR \cdot L_R}{dS_R \cdot L_R + \frac{dS_{ext}}{\mathcal{L} \cdot L_S \cdot \| \mathcal{D} \|}}
\end{equation}
\textbf{Engineering Logic:} This tier explicitly accounts for internal and external ``friction'' through the \textbf{Disorder Coefficients} ($L_R$ and $L_S$), which quantify the sequestration of logical action established in Section~\ref{sec:formal_theory}:

\begin{itemize}
    \item \textbf{Rule Disorder ($L_R$):} Defined as the ratio of the element-wise absolute path to the net vector displacement ($\|\sum |\Delta \theta|\| / \|\sum \Delta \theta\|$). High $L_R$ indicates the model is vibrating intensely in parameter space without achieving net structural advancement. Since internal chaos increases the effective rule-density, both $dR$ and $dS_R$ are multiplied by $L_R$.
    
    \item \textbf{State Disorder ($L_S$):} Represents the loss of coherence in output expressions. A high $L_S$ signifies that environmental feedback is blurred by noise, effectively dividing and diluting the useful external gain $dS_{ext}$.
\end{itemize}

Tier 3 is uniquely capable of detecting optimization pathologies, such as ``vibrating in place,'' where high parameter-space volatility ($L_R \uparrow$) signals that the agent's logical action is being entirely consumed by internal friction. By standardizing these metrics through $\| \mathcal{D} \|$ and scaling them by the effective information density ($\mathcal{L}^{-1}$), Tier 3 provides the most precise assessment of the system's proximity to the \textbf{computational wall}.

\subsection{Inertia-Aware Regulation and Learning Rate Contraction}
\label{sec:regulation_protocol}

Once the system velocity $v$ is measured, it functions as the primary feedback signal for a protective regulation protocol. The core of this mechanism is the \textbf{Relativistic Contraction of the Learning Rate}, which ensures that the agent's evolutionary tempo remains synchronized with its internal physical limits.

\subsubsection*{The Physical Nature of Learning Rate and Entropy Expulsion}

In the engineering realization of intelligence dynamics, the learning rate $\eta$ is formally identified as the projection of the system's \textbf{Characteristic Cycle ($l$)}—the internal logical clock—onto the training epoch timeline. As established in our micro-physical model (Section~\ref{sec:micro_physical_model}), $l$ defines the logical window required to maintain causal consistency during structural evolution. Consequently, $\eta$ represents the characteristic scale at which the agent samples, filters, and integrates environmental information into its rule-set \cite{kingma2014adam}.

Physically, this integration process is governed by the system's capacity for \textbf{entropy expulsion} (manifesting empirically in neural networks as the reduction of gradient variance and the smoothing of residual errors). Unlike traditional Information Bottleneck theories that focus solely on the reduction of internal representation entropy \cite{shwartz2017opening}, our framework addresses the \textbf{dissipation} of that entropy into the environment. The erasure of information (learning) requires internal entropy reduction to be dissipated as heat through the diathermal boundaries of the logical container. In a neural network, this ``informational heat'' manifests as the stochastic noise and residual error generated during rule-reconfiguration.

The constraint arises from the \textbf{geometric occlusion of the cooling channels}: as the \textbf{Rule Density ($\rho$)} increases, the proportion of ``adiabatic'' surface area—dedicated to internal rule-maintenance—grows, causing the effective heat-exchange cross-section for entropy expulsion to shrink according to $\sqrt{1-v^2}$. Just as the particles in Section~\ref{sec:micro_physical_model} require a higher external intensity to find a rare diathermal exit, a high-velocity (dense) neural network faces a \textbf{``clogging effect''} where the channels for informational noise are nearly sealed. If the learning rate $\eta$—the rate at which new structural changes are forced—exceeds this shrinking expulsion capacity, the residual entropy cannot be dissipated. This accumulated ``heat'' triggers chaotic fluctuations that physically \textbf{``shatter''} existing causal rules, providing a first-principles explanation for catastrophic forgetting and the explosive instability observed in dense models \cite{kirkpatrick2017overcoming, french1999catastrophic}.

\subsubsection*{The Relativistic Brake}

To protect the agent's structural integrity, the effective learning rate must contract to match the system's real-time entropy expulsion limit. Following the Lorentz symmetry derived on the R-S Manifold, we define the \textbf{Effective Learning Rate ($\eta_{\text{eff}}$)} as:
\begin{equation}
\eta_{\text{eff}} = \eta_{\text{base}} \cdot \frac{\sqrt{1 - v^2}}{\sqrt{1 - v_{\text{base}}^2}}
\label{eq:relativistic_brake}
\end{equation}
where $\eta_{\text{base}}$ is the standard step-size determined by the optimizer, and $v_{\text{base}}$ represents the velocity recorded during the warmup calibration in Section~\ref{sec:calibration} (typically $v \approx 0.5$). 

\begin{itemize}
    \item \textbf{Logical Freezing:} As the velocity $v$ approaches the saturation limit of $1.0$, the term $\sqrt{1-v^2}$ converges to zero, causing the effective learning rate to vanish. This induces a state of \textbf{``Logical Freezing,''} a self-preservation mode where the system locks its current parameters to prevent reconfigurations that would be impossible to dissipate or interpret.
    \item \textbf{Dynamic Adaptation:} Unlike heuristic decay schedules such as cosine annealing \cite{loshchilov2016sgdr}, this contraction is a direct physical response to the agent's internal \textbf{Inertia Expansion}. When the model encounters high-inertia data that drives $v \to 1$, the system automatically decelerates to prevent structural collapse.
\end{itemize}

By implementing this ``Inertia-Aware'' brake, we guarantee that the agent's trajectory through parameter space remains within the stable bounds of the Minkowski manifold, ensuring that rule-changes remain anchored to the system's interpretability scale $\mathcal{D}$.

\subsection{Directional Coherence and Phase Alignment}
\label{sec:direction_coherence}

While the magnitude of velocity $v$ determines the required Lorentzian contraction of the learning rate, its \textbf{direction}—defined as the orientation of the velocity vector $\vec{v}$ within the tangent space of the R-S Manifold—dictates the geometric validity of the structural evolution \cite{dauphin2014identifying}.

\subsubsection*{The Physical Meaning of Velocity Direction}
In the phase space of an agent, the orientation of $\vec{v}$ represents the \textbf{Logical Orientation} of reconfiguration, identifying which cognitive pathways or rule-subsets are being prioritized for modification. 
\begin{itemize}
    \item \textbf{Directional Coherence}: When the current trajectory aligns with historical pathways that have demonstrated stable entropy expulsion, the system performs a ``coherent'' update that preserves the agent's established topological structure \cite{fort2019emergence}.
    \item \textbf{Directional Dissonance}: An abrupt shift in orientation (e.g., updates becoming orthogonal to established gradients) indicates a phase mismatch. This suggests the system is forcing a reconfiguration that contradicts its \textbf{ontological support}, leading to a high risk of \textbf{structural shattering}.
\end{itemize}

\subsubsection*{Coherent State Anchors and Component Metrics}
To quantify this alignment, the regulator utilizes \textbf{Coherent State Anchors}— reference vectors captured during high-efficiency learning phases where entropy expulsion was optimal. We define the \textbf{Phase Coherence Score ($C$)} by evaluating the similarity between the current dynamics and these anchors across three key dimensions:

1. \textbf{Internal Directional Alignment ($C_{S_R}$)}: The cosine similarity between the current internal state displacement vector and the anchor's reference displacement:
\begin{equation}
    C_{S_R} = \cos(\vec{dS}_{R, curr}, \vec{dS}_{R, anchor}) = \frac{\vec{dS}_{R, curr} \cdot \vec{dS}_{R, anchor}}{\|\vec{dS}_{R, curr}\| \|\vec{dS}_{R, anchor}\|}
\end{equation}

2. \textbf{External Response Alignment ($C_{S_{ext}}$)}: The cosine similarity between the current external gain vector (causal ripple) and the anchor's reference gain:
\begin{equation}
    C_{S_{ext}} = \cos(\vec{dS}_{ext, curr}, \vec{dS}_{ext, anchor})
\end{equation}

3. \textbf{Efficiency Ratio Consistency ($C_{\phi}$)}: This metric evaluates whether the ``gearing'' of the update matches the anchor's optimal efficiency. Let $r = \|\vec{dS}_{ext}\| / \|\vec{dS}_R\|$. We define the ratio consistency coefficient $k_{ratio}$ as:
\begin{equation}
    k_{ratio} = \frac{\min(r_{curr}, r_{anchor})}{\max(r_{curr}, r_{anchor})}
\end{equation}
The resulting consistency score is scaled via a sine function to ensure a smooth penalty:
\begin{equation}
    C_{\phi} = \sin\left(\frac{\pi}{2} \cdot k_{ratio}\right)
\end{equation}

The total \textbf{Phase Coherence Score ($C$)} is derived from the product of these metrics: 
\begin{equation}
    C = C_{S_R} \cdot C_{S_{ext}} \cdot C_{\phi}
\end{equation}
This score acts as a final multiplicative gate for the learning rate, ensuring that updates are only permitted when the system is both below the velocity saturation limit and geometrically aligned with a stable history:
\begin{equation}
    \eta_{final} = \eta_{eff} \cdot C
\end{equation}
It should be noted that while this specific vector-calculus framework offers a robust engineering realization, it is \textbf{not the exclusive method} for implementing intelligence inertia dynamics. Besides, Several high-dimensional phenomena must be addressed for practical deployment. 
First, the \textbf{intrinsic sparsity} of neural gradients can bias directional metrics; this is mitigated by performing alignment in effective subspaces or utilizing projection masks. 
Second, \textbf{stochastic noise} becomes dominant as $v \to 1$ and the external signal $dS_{\text{ext}}$ diminishes, necessitating low-pass filtering or sliding-window integration to distinguish structural density from transient fluctuations. 
Finally, to address the \textbf{stability-plasticity dilemma} \cite{parisi2019continual}, a \textbf{logical unfreezing mechanism}—facilitated by a spontaneous exponential decay of measured inertia—is required to release the relativistic brake once the core structure has stabilized, allowing the agent to recover the plasticity necessary for new information.

\section{Experiments}
\label{sec:experiments}

In this chapter, we subject the derived \textbf{physical principles} of \textit{intelligence inertia} to a series of rigorous empirical investigations. Transitioning from the micro-physical derivations in Section~\ref{sec:micro_physical_model}, the formal axiomatization in Section~\ref{sec:formal_theory}, and the engineering realization in Section~\ref{sec:engineering_realization}, we demonstrate how these abstract laws manifest within the actual tensor dynamics of deep neural networks \cite{lecun2015deep}. The experimental suite utilizes established architectures, such as ResNet-18 \cite{he2016deep}, and standard benchmarks like CIFAR-10 \cite{krizhevsky2009learning}, and is organized into three logically advancing stages designed to validate the framework's predictive power and engineering utility:

\begin{enumerate}
    \item \textbf{Experiment I: Decisive Adjudication of Intelligence Inertia Divergence.} 
    This stage aims to confirm the physical reality of the \textbf{Informational Speed Limit} within the \textbf{R-S Manifold} and the resulting \textbf{Inertia Expansion} effect. By simulating high-velocity regimes characterized by a vanishing external gain ($dS_{\text{ext}} \to 0$), we observe whether computational work follows the non-linear, relativistic trajectories predicted by our framework, thereby establishing the empirical bedrock of the theory.

    \item \textbf{Experiment II: Evolutionary Geometry and the Reachability Topography.} 
    We investigate the fundamental impact of architectural optimization on system inertia. By mapping the performance landscape across various neural topologies within the phase space of the \textbf{R-S Manifold}, we verify that maintaining a balanced velocity of $v \approx 0.5$—the ``golden axis'' of energy equipartition—serves as the steepest descent path for intelligent evolution, providing a principled physical methodology for architectural design.

    \item \textbf{Experiment III: Engineering Practice --- The Inertia-Aware Scheduler Wrapper.} 
    We deploy the practical \textbf{Inertia-Aware Scheduler Wrapper} based on the implementation protocols defined in Section~\ref{sec:engineering_realization}. Through evaluations of dynamic performance, resilience against high-entropy logic shocks (noise), and memory retention in continual learning \cite{kirkpatrick2017overcoming}, we demonstrate the superior efficiency and stability of systems that respect their intrinsic physical resistance to change.
\end{enumerate}

\subsection{Experiment I: Decisive Adjudication of Intelligence Inertia Divergence}

\label{sec:exp_adjudication}
The primary objective of this experiment is to provide a decisive empirical test between classical information-geometric models and our relativistic framework of \textbf{Intelligence Inertia}. By subjecting a neural network to extreme logical stress, we aim to observe whether the computational work required for adaptation remains a quadratic function of velocity—as predicted by the \textbf{Fisher Information Matrix (FIM)} baseline—or whether it exhibits the non-linear \textbf{relativistic $J$-shaped inflation curve} characteristic of Inertia Expansion.

\subsubsection{Experimental Design and Physical Mapping}

To empirically validate the existence of the ``computational wall,'' we constructed an environment that forces an agent through a spectrum of rule densities, ranging from the low-velocity ``clean data'' regime to the high-velocity ``informational limit.'' This was achieved by systematically injecting label noise (ranging from 0\% to 100\%) into the CIFAR-10 dataset.

In classical machine learning, the injection of label noise is known to induce a highly non-convex and rugged loss landscape, stripping the gradient of clear directional signals \cite{keskar2016large}. From our physical perspective, this widely observed phenomenon maps isomorphically to the suppression of the external gain ($dS_{ext} \to 0$); random labels provide no coherent structure for the model to project onto the state manifold. Consequently, to minimize the training objective within this chaotic environment, the system is forced to undergo intense \textbf{Internal Reconfiguration ($dS_R \to \infty$)} to memorize the noise, thereby pushing the local rule density ($v = \rho$) toward its saturation limit. We utilized the ResNet-18 architecture with 11.2M (Million parameters) as our physical substrate, recording the total \textbf{computational work ($W$)}, measured in epochs, required to reach a specific convergence threshold across different noise energy levels.

The dynamical parameters are measured according to the engineering protocols established in Tier 2, Section~\ref{sec:implementation_tiers}. The adjudication logic directly tests the validity of the metric tensor probe defined in Eq. (\ref{eq:normalized_velocity}). By manipulating the environmental feedback ($dS_{ext} \to 0$), we actively drive the independent variable---the local velocity $v$---across the $(0, 1)$ spectrum. The critical observation is then directed at the dependent physical variable: the total computational work. If the classical FIM framework is exhaustive, the measured computational cost should follow a smooth, quadratic growth ($v^2$) indicative of local stiffness. Conversely, if the Intelligence Inertia framework holds true, the empirically measured work must precisely track the non-linear Lorentzian divergence ($\gamma - 1$) as $v$ approaches the informational speed limit, confirming dynamic mass expansion over static geometric curvature.

\subsubsection{Dynamical Models and Regression Basis}

To adjudicate the empirical results, we establish two competing mathematical models for regression analysis:

1. \textbf{Classical Dynamical Hypothesis (FIM Baseline):} This model assumes that the learning cost is a direct function of the local manifold curvature, following a second-order Taylor expansion analogous to Newtonian kinetic energy:
\begin{equation}
    Cost_{classical} = k \cdot v_{rel}^2 + b
\end{equation}

2. \textbf{Relativistic Dynamical Hypothesis (Intelligence Inertia Model):} This model assumes that the system’s \textbf{Effective Mass} undergoes a geometric expansion as velocity approaches the informational horizon (where $c = \rho_{\max} = 1$):
\begin{equation}
    Cost_{relativistic} = k \cdot (\gamma - 1) + b = k \cdot \left( \frac{1}{\sqrt{1 - (v_{rel}/c)^2}} - 1 \right) + b
\end{equation}

By fitting these two equations to the observed computational expenditure across the noise-injected velocity spectrum, we can determine whether the agent's resistance to change is a static property of curvature or a dynamic consequence of relativistic mass divergence.
\subsubsection{Experimental Results and Attribution Analysis}

Following the execution of our controlled noise-injection protocol, we evaluated the predictive performance of the competing dynamical models. The resulting data provides a stark contrast between the local approximations of classical information geometry and the global consistency of the \textbf{Intelligence Inertia} framework. The quantitative results of this adjudication are summarized in Table~\ref{tab:model_comparison}.

\begin{table}[htbp]
\centering
\caption{Dynamical Model Fitting Performance across Coordinate Systems. The Intelligence Inertia Theory, utilizing relativistic mass expansion, consistently outperforms classical Fisher Information approximations, particularly as the system approaches the informational speed limit.}
\label{tab:model_comparison}
\begin{tabular}{@{} m{0.22\textwidth} m{0.22\textwidth} m{0.12\textwidth} m{0.38\textwidth} @{}}
\toprule
\centering \textbf{Dynamical Model} & \centering \textbf{Reference Frame Treatment} & \centering \textbf{Fit Error (RMSE)} & \centering \textbf{Performance} \tabularnewline
\midrule
\centering Classical FIM & \centering Absolute Coordinates & \centering 36.0 & \textbf{Failed}: Incapable of capturing the non-linear ``J-Curve'' divergence. \\ 
\centering Classical FIM & \centering Galilean Shift ($v - v_0$) & \centering 30.0 & \textbf{Limited}: Fits the low-speed regime but severely underestimates the wall effect. \\ 
\centering Hybrid FIM & \centering Lorentz Transformation & \centering 25.5 & \textbf{Suboptimal}: Corrects for velocity addition but lacks mass expansion logic. \\ 
\centering \textbf{Relativistic Mass (Intelligence Inertia)} & \centering \textbf{Universal Covariance} & \centering \textbf{18.5 – 19.6} & \textbf{Optimal}: Precisely fits the asymptotic divergence and exhibits frame independence. \\ \bottomrule
\end{tabular}
\vspace{2pt}
\end{table}

\textbf{Experimental Conclusion}: Experiment I definitively confirms the existence of a fundamental \textbf{Rule-Density Limit} in intelligent agents, functioning as an informational ``speed of light'' ($c \approx 1$). This validates the micro-physical derivation in Section~\ref{sec:micro_physical_model}, proving that the resistance to change is not merely a consequence of manifold curvature but a result of the system's logical budget reaching saturation. While the classical Fisher Information Matrix (FIM) serves as a reasonable second-order approximation in low-velocity states, \textbf{Inertia Expansion} becomes the dominant physical factor governing computational work as the system nears its cognitive horizon.

To further isolate the causative mechanisms, we performed a structural attribution analysis using a four-dimensional arena comparison \cite{myung2000importance}. This methodology allows us to rigorously adjudicate between competing hypotheses by testing the necessity of mass expansion and the robustness of the theory across different observational reference frames.
\paragraph{1. Arena I \& II: Reference Frame Sensitivity and Model Robustness}

The first stage of our attribution analysis investigates the dependency of each model on the selection of the coordinate origin—specifically, whether the model's validity relies on the arbitrary subtraction of the system's initial ``rest velocity'' ($v_0$). This tests whether the observed dynamics are a fundamental feature of the R-S Manifold or merely a consequence of specific observational framing.

\begin{figure}[htbp]
\centering
\includegraphics[width=\textwidth]{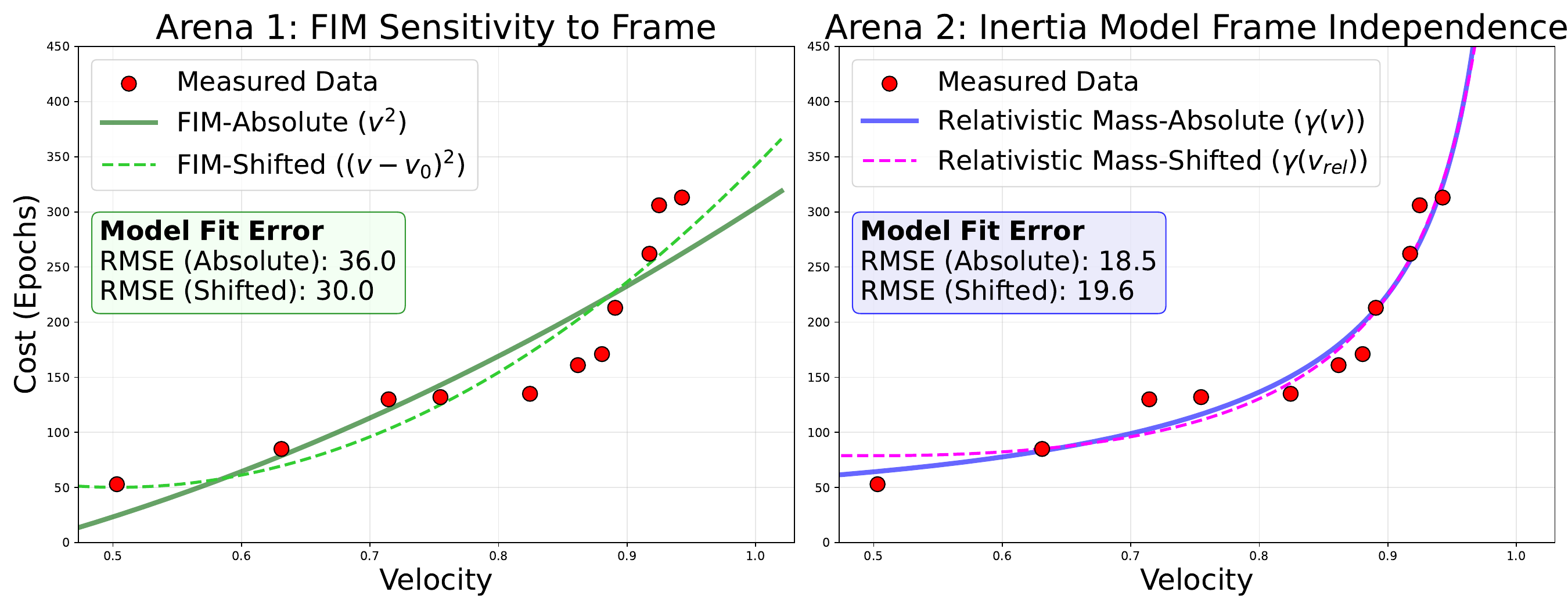}
\caption{Comparative Adjudication of Reference Frame Sensitivity and Model Robustness. Arena 1 \textbf{(Left)} demonstrates the sensitivity of the classical FIM model to coordinate shifts; the model fails to track the data trend in absolute coordinates. Arena 2 \textbf{(Right)} illustrates the Intelligence Inertia model, where the relativistic curves remain covariant and highly accurate regardless of the reference frame, maintaining an RMSE between 18.5 and 19.6.}
\label{fig:reference_frame_arena}
\vspace{2pt}
\end{figure}

\begin{itemize}
    \item \textbf{Reference Frame Dependency of FIM (Arena 1)}: 
    In the \textbf{Absolute Reference Frame} (without subtracting $v_0$), the FIM prediction fails catastrophically (RMSE = 36.0), appearing as a linear approximation that cannot capture the rising work-divergence. After applying a \textbf{Galilean Transformation} to account for $v_0$ (FIM-Shifted), the error drops to 30.0. However, while this improves the fit in the low-velocity regime, it remains blind to the explosive \textbf{Inertia Expansion} encountered when $v > 0.9$. This confirms that FIM acts only as a local, second-order approximation valid at the origin \cite{martens2020optimizing}.
    
    \item \textbf{Lorentz Covariance of Intelligence Inertia Model (Arena 2)}: 
    The relativistic framework demonstrates profound \textbf{Frame Independence}. Unlike classical approximations, the predictive law exhibits \textbf{mathematical covariance}; whether evaluated in absolute or shifted coordinates, the model consistently captures the intrinsic hyperbolic geometry of the \textbf{Inertia Expansion} (RMSE 18.5--19.6). This confirms that the observed ``Computational Wall'' is an \textbf{invariant property} of the R-S Manifold rather than an artifact of coordinate selection \cite{weinberg1972gravitation}. 
\end{itemize}

\textbf{Physical Conclusion}: These results establish that \textbf{Intelligence Inertia ($\mu$)} is not a mere empirical heuristic dependent on measurement methods, but an \textbf{intrinsic physical law} that describes the fundamental cost of structural evolution in intelligent systems.

\paragraph{2. Arena III \& IV: Necessity of Velocity Transformation vs. Mass Expansion}

The second stage of our analysis utilizes an ``ablation study'' approach to disentangle the specific contributions of relativistic kinematics (velocity definitions) from the dynamics of \textbf{Inertia Expansion}. This comparison determines whether the observed non-linearity is a result of coordinate transformation logic or a fundamental shift in the system's physical resistance as it approaches the \textbf{Interpretability Criterion} limit ($ds^2 \to 0$).

\begin{figure}[htbp]
\centering
\includegraphics[width=\textwidth]{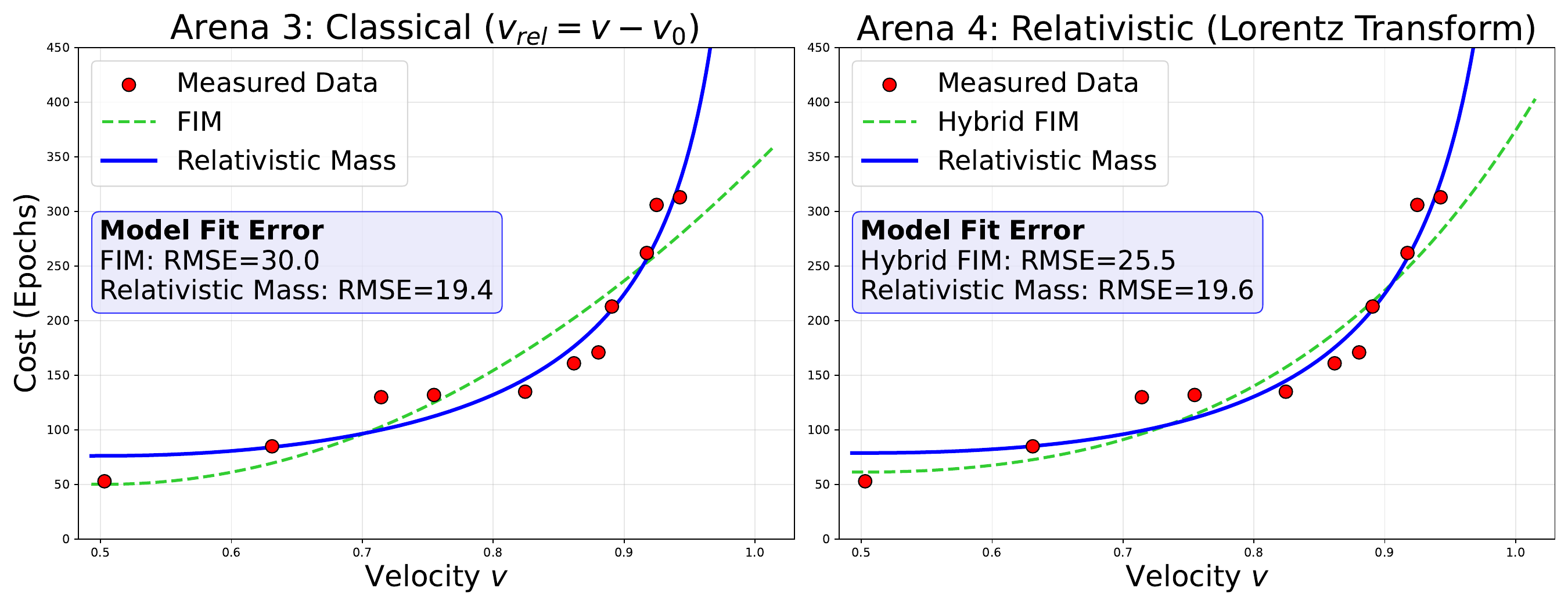}
\caption{\textbf{Ablation Analysis of Relativistic Velocity Addition vs. Mass Expansion.} Arena 3 \textbf{(Left)} contrasts the Galilean-shifted FIM against the relativistic mass model, showing the clear failure of the quadratic assumption at high speeds; Arena 4 \textbf{(Right)} introduces a ``Hybrid FIM'' model, which applies the relativistic Lorentz velocity transformation but retains the classical quadratic cost formula, highlighting that velocity correction alone is insufficient to explain the ``J-Curve'' inflation.}
\label{fig:arena_3_4}
\end{figure}

\begin{itemize}
    \item \textbf{Limitations of Velocity Transformation (Hybrid FIM):} To test if the error stemmed solely from velocity measurement, we constructed a \textbf{Hybrid FIM} model. This model utilizes the relativistic velocity addition rule to correct the system's speed but maintains the classical quadratic work function. The results indicate that even with the corrected velocity, the Hybrid FIM (\textbf{RMSE = 25.5}) remains incapable of explaining the vertical surge in computational work at the high-velocity limit. This suggests that the non-linearity is not a kinematic artifact but a dynamical property of sequestered energy.
    \item \textbf{Decisiveness of Inertia Expansion:} The complete Intelligence Inertia model, which fully incorporates the $\gamma$ factor (\textbf{RMSE = 19.6}), precisely captures the ``computational wall'' encountered as the agent approaches the informational speed limit. This confirms that the sequestration of logical action leads to an actual inflation of the system's effective mass.
\end{itemize}

    \textbf{Physical Conclusion:} These results provide a robust empirical refutation of the hypothesis that observed cost anomalies are merely artifacts of linear velocity errors. While classical optimization theory qualitatively attributes the difficulty of noise-fitting to the non-convexity of the loss landscape, our framework provides the exact \textbf{quantitative metric tensor explanation}. The divergence of computational work at the reachability boundary is fundamentally governed by the \textbf{relativistic divergence of the system's effective mass} ($\gamma$) during intensive Internal Reconfiguration. Thus, the Intelligence Inertia framework translates the qualitative "ruggedness" of a loss basin into a precise mathematical prediction of macroscopic computational expenditure.

\subsection{Experiment II: Evolutionary Geometry and the Reachability Topography} 
\label{sec:exp_evolutionary_geometry}

The empirical confirmation of the Inertia Expansion effect in Experiment~\ref{sec:exp_adjudication} establishes that the effective mass of an agent diverges as its velocity $v$ approaches the informational speed limit. This physical reality shifts the focus of architectural engineering from merely increasing parameter depth to \textbf{minimizing the work performed against internal inertia} during the learning process. Within this stage of our inquiry, we map the ``Reachability Topography'' of neural architectures on the R-S Manifold to determine how topological choices influence the coordination between Internal Reconfiguration ($dS_R$) and environmental feedback.

\subsubsection{Experimental Objective: From Inertia Evasion to Dynamical Optimization} 
\label{sec:exp_obj_optimization}

While traditional theories based on \textbf{Hausdorff dimensions} suggest that increasing architectural complexity should exponentially reduce learning costs \cite{falconer2014fractal}, the \textbf{Intelligence Inertia Theory} posits that efficiency is not a simple function of dimensionality \cite{kaplan2020scaling}. Instead, it is constrained by the coordination between the internal state shift driven by rule reconfiguration ($dS_R$) and the external state gain provided by the environment ($dS_{ext}$), as governed by the local velocity equation:
\begin{equation}
    v = \rho = \frac{dR}{dS_R + dS_{ext}}
\end{equation}
This framework suggests that architectural design is essentially a problem of \textbf{dynamical balancing}. The objective of this experiment is to identify the optimal ``geodesic'' for structural progress by testing three specific predictions:

\begin{enumerate}
    \item \textbf{Limitations of Single-Axis Optimization}: Improving an architecture along only one dimension—either by smoothing the internal manifold (optimizing $dS_R$, e.g., via residual connections \cite{he2016deep}) or by increasing inductive biases (optimizing $dS_{ext}$, e.g., via multi-scale features)—will fail to achieve global energy minimality. Such systems inevitably encounter a bottleneck of diminishing returns as their Rule Density deviates from the optimal regime.
    \item \textbf{Orthogonal Synchronous Evolution}: We hypothesize that the most efficient evolutionary path requires internal effort and external feedback to be optimized in synchronization. This keeps the system velocity dynamically anchored near $v \approx 0.5$, the point of \textbf{Energy Equipartition} \cite{huang1987statistical} (a balanced allocation of computational budget between internal weight reconfiguration and external representation learning), where the resistance to structural change is minimized.
    \item \textbf{Saddle-Shaped Decay Surface}: Macroscopically, while learning costs decay as architectural rules improve, this decay is non-linear. We predict the cost surface on the R-S Manifold will exhibit a characteristic \textbf{Saddle} geometry \cite{dauphin2014identifying}, where the steepest descent is achieved through a ``Zig-Zag'' path that alternates between rule-refinement and state-expansion.
\end{enumerate}

\subsubsection{Experimental Design and Quantitative Measurement}
\label{subsec:exp_design_quant}

To isolate the impact of architectural topology on system inertia, we strictly constrained the parameter budget of all candidate architectures to 5.0M and fixed the network depth to 10 layers. Within this framework, we utilize the \textbf{Reachability Limit ($\mathcal{L}_{min}$)}—defined as the minimum achievable loss on the CIFAR-10 task \cite{krizhevsky2009learning}—as the primary indicator of the physical ``work cost'' required to reach a specific intelligence state. Total floating-point operations (TFLOPs) serve as a secondary measure of the actual computational energy expended during the process.

We identify two orthogonal dimensions of architectural evolution on the \textbf{R-S Manifold} that govern the system's dynamical state:
\begin{enumerate}
    \item \textbf{The Internal Rule-Reconfiguration ($dS_R$) Axis}: This axis represents architectural enhancements designed to smooth the internal parameter manifold and reduce frictional losses during structural updates. The progression includes the transition from \textbf{Baseline (R0)} to \textbf{Batch Normalization (BN, R1)} \cite{ioffe2015batch} and \textbf{Residual Connections (Res, R2)}, aiming to minimize the internal effort required for a given logic shift.
    \item \textbf{The External State Gain ($dS_{ext}$) Axis}: This axis represents the integration of inductive biases \cite{mitchell1980need} that allow the system to effectively align with the environmental manifold. Improvements progress from \textbf{Baseline (S0)} to \textbf{Locality-based Convolutional Neural Networks (CNN, S1)} \cite{lecun1998gradient} and \textbf{Multi-scale Convolutional Neural Networks (MCNN, S2)} \cite{szegedy2015going}, increasing the productive state-shift obtained from each rule-update.
\end{enumerate}

The Multi-Layer Perceptron (MLP) \cite{rosenblatt1958perceptron} is established as our \textbf{Evolutionary Origin (R0, S0)}. As a structure with minimal inductive bias and baseline connectivity, it represents a state of ``maximal ignorance'' within the theory. This origin is used to calibrate the \textbf{Symbolic Granularity ($\mathcal{D}$)}, anchoring the system’s initial velocity at the point of Energy Equipartition ($v \approx 0.5$). 

The core dynamical parameters captured across the $3 \times 3$ architectural matrix are presented in Table~\ref{tab:arch_comparison}.

\begin{table}[ht!]
\centering
\caption{Comparative Analysis of Dynamical Parameters and Reachability Limits across Neural Architectures. Abbreviations for architectures: MLP (Multi-Layer Perceptron), BN (Batch Normalization), Res (Residual), CNN (Convolutional Neural Network), MCNN (Multi-scale Convolutional Neural Network). Column Abbreviations: Rule Reconfig. (Internal Rule-Reconfiguration), Ext. Gain (External State Gain), Vel. (Velocity), Vel. Dev. (Velocity Deviation from 0.5), Reach. Limit (Reachability Limit $\mathcal{L}_{min}$), Work (Computational Effort in TFLOPs). The data shows that the lowest reachability limit is attained by architectures like Res-MCNN, which achieve optimal balance between rule effort and state feedback, minimizing the deviation from the $v \approx 0.5$ golden axis.}
\label{tab:arch_comparison}
\begin{tabular}{>{\centering\arraybackslash}m{0.11\textwidth}>{\centering\arraybackslash}m{0.14\textwidth}>{\centering\arraybackslash}m{0.14\textwidth}>{\centering\arraybackslash}m{0.06\textwidth}>{\centering\arraybackslash}m{0.1\textwidth}>{\centering\arraybackslash}m{0.1\textwidth}>{\centering\arraybackslash}m{0.1\textwidth}}
\toprule
\textbf{Arch. Name} & \textbf{Rule Reconfig. ($dS_R$ Axis)} & \textbf{Ext. Gain ($dS_{ext}$ Axis)} & \textbf{Vel. ($v$)} & \textbf{Vel. Dev. $|v - 0.5|$} & \textbf{Reach. Limit ($\mathcal{L}_{min}$)} & \textbf{Work (TFLOPs)} \tabularnewline
\midrule
\textbf{MLP} (Origin) & Baseline (R0) & Baseline (S0) & 0.502 & 0.002 & 2.302 & 0.6 \tabularnewline
\textbf{BN} & Normalized (R1) & Baseline (S0) & 0.522 & 0.022 & 1.389 & 68.7 \tabularnewline
\textbf{Res} & Residual (R2) & Baseline (S0) & 0.818 & 0.318 & 1.322 & 117.0 \tabularnewline
\textbf{MLP-CNN} & Baseline (R0) & Locality (S1) & 0.179 & 0.321 & 1.485 & 830.0 \tabularnewline
\textbf{MLP-MCNN} & Baseline (R0) & Multi-scale (S2) & 0.194 & 0.306 & 1.339 & 3670.0 \tabularnewline
\textbf{BN-CNN} & Normalized (R1) & Locality (S1) & 0.532 & 0.032 & 0.647 & 664.0 \tabularnewline
\textbf{Res-CNN} & Residual (R2) & Locality (S1) & 0.689 & 0.189 & 0.578 & 863.0 \tabularnewline
\textbf{BN-MCNN} & Normalized (R1) & Multi-scale (S2) & 0.531 & 0.031 & 0.599 & 2480.0 \tabularnewline
\textbf{Res-MCNN} & Residual (R2) & Multi-scale (S2) & 0.452 & 0.048 & \textbf{0.553} & 2090.0 \tabularnewline
\bottomrule
\end{tabular}
\end{table}

\subsubsection{Results Analysis: Reachability Topography and the Valley of Equilibrium}

The experimental results from the $3 \times 3$ architecture matrix reveal a profound geometric structure in the cost landscape of neural evolution. By plotting the Reachability Limit ($\mathcal{L}_{min}$) against the orthogonal axes of Internal Reconfiguration ($dS_R$) and External State Gain ($dS_{ext}$) on the R-S Manifold, we observe the emergence of a characteristic \textbf{Saddle Topography} \cite{li2018visualizing}.

\paragraph{1. The Saddle Geometry of Reachability}

\begin{figure}[ht]
    \centering
    \includegraphics[width=1\textwidth]{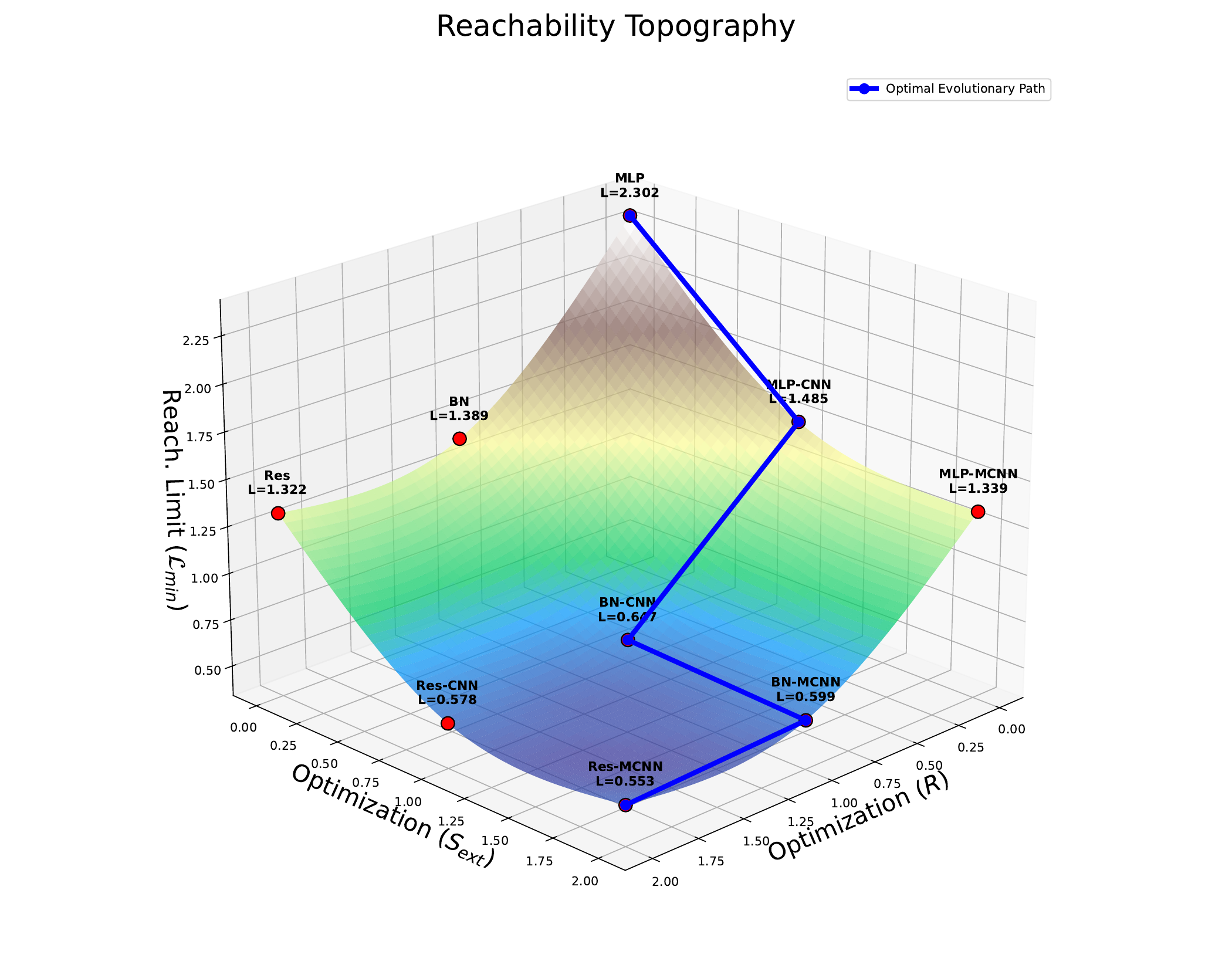}
    \caption{3D Reachability Topography and the Zig-Zag Evolutionary Geodesic. This 3D surface plots the Reachability Limit ($\mathcal{L}_{min}$) as a function of improvements in internal manifold smoothing ($dS_R$-axis) and external structural bias ($dS_{ext}$-axis). The red nodes represent measured architectures from our matrix, while the blue line illustrates the ``Zig-Zag Geodesic,'' representing the path of maximum efficiency. The topography exhibits a distinct saddle shape, where the steepest descent occurs along the diagonal of balanced development.}
    \label{fig:reachability_topography}
\end{figure}

Figure~\ref{fig:reachability_topography} visualizes the minimum achievable loss across the evolutionary surface, illustrating how different structural combinations influence the agent's ultimate intelligence capacity.

The analysis of this topography yields three critical insights into the dynamics of intelligence:

\begin{itemize}
    \item \textbf{Plateaus in Axial Trajectories}: Observing the \textbf{MLP $\to$ Res} path (pure $dS_R$ optimization), we find that while manifold smoothing provides an initial gain, $\mathcal{L}_{min}$ quickly hits a plateau, decreasing only from 2.302 to 1.322 before the slope flattens. A similar diminishing return is observed in the \textbf{MLP $\to$ MLP-MCNN} path (pure $dS_{ext}$ optimization). This confirms that optimizing only one dimension of the Rule-State pair—regardless of the sophistication of the technique—leads the system to deviate from the golden axis, resulting in an \textbf{Inertia Expansion} that halts further progress.
    \item \textbf{Synergistic Jumps via Orthogonal Transitions}: The most significant performance leaps occur at orthogonal switching points, such as the transition from \textbf{MLP-CNN to BN-CNN}. These ``jumps'' indicate that when a system is stalled on an axial plateau, the introduction of a complementary structural dimension restores the dynamical balance, allowing for a rapid collapse in total inertia and a corresponding drop in reachable loss.
    \item \textbf{The Saddle Geometry and the Geodesic}: The topography clearly demonstrates that the ``bottom'' of the informational valley follows a diagonal trajectory toward the (2,2) coordinate. This \textbf{Zig-Zag Geodesic} represents the sequence of modifications where internal effort and external feedback remain in constant proportion. This validates the \textbf{Law of Orthogonal Synergy}: architectural progress is maximized only when rule reconfiguration and state expression are improved in synchronization, keeping the system velocity dynamically anchored within the \textbf{Valley of Equilibrium}.
\end{itemize}

\paragraph{2. The Valley of Dynamic Equilibrium}
To identify the underlying optimality of the Zig-Zag path, we analyzed the \textbf{Velocity Deviation ($|v - 0.5|$)} across the architectural landscape, representing the system's distance from the theoretical point of \textbf{Energy Equipartition} \cite{huang1987statistical}.

\begin{figure}[htb]
    \centering
    \includegraphics[width=0.8\textwidth]{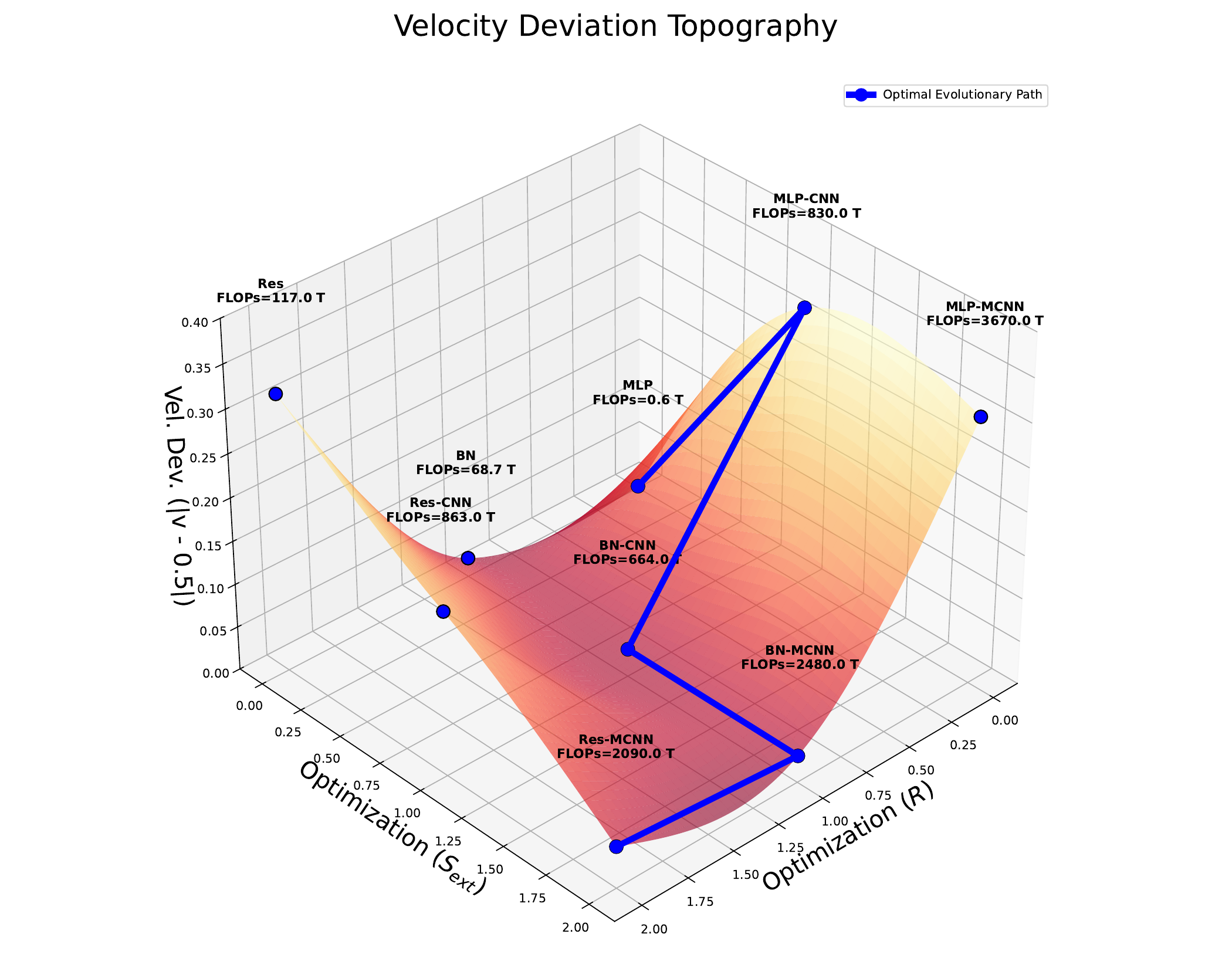}
    \caption{Velocity Deviation Topography and the Dynamical Riverbed. This 3D surface maps the absolute deviation of system velocity from the $0.5$ golden axis. The resulting ``V-shaped'' valley—the Dynamical Riverbed—approximately aligns with the optimal evolutionary path, identifying the minimization of velocity deviation as the physical objective of architectural progress.}
    \label{fig:velocity_deviation}
\end{figure}

\begin{itemize}
    \item \textbf{The Dynamical Riverbed}: Figure~\ref{fig:velocity_deviation} reveals a striking ``Riverbed'' geometry on the R-S Manifold. We observe that the observed optimal evolutionary trajectory \textbf{approximates the trough of this valley}. This confirms that superior architectural reachability is achieved not by singular scaling, but by the continuous balancing of Internal Reconfiguration and External State Gain, effectively maintaining the system near the state of energy equipartition where resistance to change is minimized.
    \item \textbf{Efficiency Adjudication}: The topography explains the suboptimal performance of unbalanced architectures. MLP-CNN ($v = 0.179$) represents an ``environmental stall'' where high external state gain without sufficient internal rule-reconfiguration capacity drives the system toward the $v \to 0$ limit. Conversely, the Res architecture ($v = 0.818$) exhibits an Inertia Expansion, where excessive rule-flexibility without sufficient environmental alignment pushes the system toward the relativistic $v \to 1$ wall. Architectures positioned near the base of the Riverbed maintain the dynamical equilibrium required for efficient learning \cite{saxe2013exact}.
    \item \textbf{Geodesic Principle}: These results suggest that architectural evolution is fundamentally a search for the geodesic—the trajectory that seeks to minimize the velocity deviation $|v - 0.5|$ to overcome the system's intrinsic Intelligence Inertia.
\end{itemize}

\subsubsection{Summary and Guidelines for Architectural Evolution}
Experiment II demonstrates that \textbf{Intelligence Inertia} is a topological property of neural networks. Even with a constant parameter count, an optimized topology like Res-MCNN can reduce the reachability residual by a factor of four compared to the baseline MLP. Our analysis establishes the \textbf{Methodology of Bipedal Evolution}: architectural progress must alternate between optimizing \textbf{Internal Reconfiguration ($dS_R$)} and improving \textbf{External State Gain ($dS_{ext}$)}. Any single-axis improvement will inevitably cause the velocity to drift from the $0.5$ axis, leading to a collision with either the environmental or relativistic ``walls.''

\subsection{Experiment III: Engineering Practice --- The Inertia-Aware Scheduler Wrapper}
\label{sec:experiment_iii}

Experiment~\ref{sec:exp_evolutionary_geometry} established that architectural progress is macroscopically driven by a trajectory toward the energy equipartition point of $v \approx 0.5$. This topological finding raises a critical microscopic hypothesis: if structural evolution inherently favors this balanced velocity, can the optimization process be enhanced by dynamically scaling the Internal Reconfiguration ($dS_R$) based on the \textbf{Lorentzian contraction} of $v$ in real-time? To investigate this, we implemented the \textbf{Inertia-Aware Scheduler Wrapper}. This engineering tool functions as a collaborative regulatory layer, nesting atop standard learning rate policies to align microscopic training dynamics with the physical principles of Intelligence Inertia \cite{li2017stochastic}.

The experiment evaluates the practical utility of the Scheduler wrapper across three distinct scenarios:
\begin{enumerate}
    \item \textbf{Convergence Limit Test}: The Scheduler wrapper is deployed as a ``physical enhancement plugin'' atop eight mainstream learning rate schedulers in a nested configuration. This quantifies its universal effectiveness in accelerating convergence and compressing the \textbf{Reachability Limit} ($\mathcal{L}_{min}$).
    \item \textbf{Noise Shock Resilience}: In a standalone configuration, we evaluate the Scheduler Wrapper’s capacity to protect the agent's structural integrity against high-entropy logic shocks, utilizing baseline algorithms only to provide a terminal learning rate boundary.
    \item \textbf{Continual Learning Stability}: We test the Scheduler Wrapper's ability to prevent \textbf{logical shattering} during abrupt task transitions without relying on replay buffers or external data caches.
\end{enumerate}

\subsubsection{Core Control Logic of the Wrapper}
\label{sec:wrapper_logic}

The Scheduler Wrapper performs a real-time audit of the agent’s evolutionary trajectory on the R-S Manifold by implementing the engineering protocols established in Section~\ref{sec:engineering_realization}. It operates through three core physical mechanisms:

\begin{enumerate}
    \item \textbf{Full-Spectrum Velocity Measurement}: The Scheduler Wrapper utilizes the \textbf{Tier 3 (Disorder-Aware)} measurement protocol defined in Section~\ref{sec:implementation_tiers}. By calculating the instantaneous rule density, it explicitly accounts for internal and external ``friction'' via the Disorder Coefficients ($L_R$ and $L_S$). This ensures that the measured velocity $v$ accurately reflects the structural stress on the manifold, distinguishing productive structural advancement from unproductive chaotic vibration.

    \item \textbf{Directional Coherence Consideration}: Following the principles of \textbf{Phase Alignment} established in Section~\ref{sec:direction_coherence}, the Scheduler wrapper audits the orientation of each velocity vector $\vec{v}$. It identifies the logical orientation of the reconfiguration; updates that align with historical coherent pathways are permitted, while those exhibiting directional dissonance (orthogonal to the established ontological support) are damped to prevent the physical shattering of existing rules.

    \item \textbf{Multiscale Geometric Coupling}: In the nested operational mode, the Scheduler wrapper utilizes a \textbf{weighted geometric mean} to integrate macroscopic scheduling with microscopic inertial protection \cite{smith2018disciplined}. The composed step-size $\eta_{composed}$ is determined as:
    \begin{equation}
        \eta_{composed} = (\eta_{sched})^{1-w} \cdot (\eta_{wrapper})^{w}
        \label{eq:geometric_coupling}
    \end{equation}
    The weight $w=0.2$ ensures that the primary scheduler ($\eta_{sched}$) dictates the overall energy decay strategy, while the physical inertia constraint ($\eta_{wrapper}$) provides a $20\%$ regulatory influence for fine-grained structural protection.
\end{enumerate}

\subsubsection{Sub-experiment I: Dynamic Performance and Reachability Limit Analysis}
\label{sec:sub_exp_1}

To evaluate the universal generalizability of the \textbf{Inertia-Aware Scheduler Wrapper}, we utilized a ResNet substrate (0.5M) and benchmarked its performance across eight mainstream scheduling algorithms provided by the PyTorch library \cite{paszke2019pytorch}. The objective of this inquiry is to observe the shift in system dynamics when traditional heuristic policies are replaced by a physical awareness of the system's trajectory on the R-S Manifold. Specifically, we measure the wrapper's ability to accelerate initial learning and compress the final \textbf{Reachability Limit ($\mathcal{L}_{min}$)}, effectively lowering the system's thermodynamic floor.

The quantitative results of this benchmark are summarized in Table~\ref{tab:scheduler_benchmark}, providing a comparative view of the ``Base'' versus ``Enhanced'' configurations. 

\begin{table}[ht!]
\centering
\caption{Comparative Statistical Analysis of Inertial Enhancement Across Schedulers. The table provides a quantitative leap in training efficiency and reachability. Sched.: PyTorch \cite{paszke2019pytorch} base scheduler, including Cosine Annealing and Cosine Restart \cite{loshchilov2016sgdr}, OneCycle Policy \cite{smith2019super}, Cyclic LR \cite{smith2017cyclical}, Multi-Step, Polynomial, Exponential, and ReduceLROnPlateau; Cfg.: Operation mode; Prog.@30: Convergence percentage at Epoch 30; Accel.: Improvement in Progress; Gain: Reduction in $\mathcal{L}_{min}$.}
\label{tab:scheduler_benchmark}
\renewcommand{\arraystretch}{1.5} 
\setlength{\tabcolsep}{5.5pt} 
\begin{tabular}{m{0.12\textwidth}<{\centering} m{0.09\textwidth}<{\centering} m{0.09\textwidth}<{\centering} m{0.1\textwidth}<{\centering} m{0.08\textwidth}<{\centering} m{0.09\textwidth}<{\centering} m{0.1\textwidth}<{\centering} m{0.09\textwidth}<{\centering}}
\toprule
Sched. & Cfg. & Loss@30 & Prog.@30 & Accel. & Reach. $\mathcal{L}_{min}$ & Gain & Best Ep. \tabularnewline
\midrule
\textbf{Pure Inertia} & Wrapper Only & \textbf{1.3392} & \textbf{94.31\%} & N/A & 1.2811 & N/A & 96 \tabularnewline
\midrule
Cosine Ann. & Base / Enh. & 1.419 / 1.395 & 82.2\% / 83.7\% & \textbf{+1.74\%} & 1.229 / 1.218 & \textbf{+0.87\%} & 98 / 97 \tabularnewline
OneCycle & Base / Enh. & 1.462 / 1.441 & 79.1\% / 80.1\% & \textbf{+1.46\%} & 1.240 / 1.227 & \textbf{+1.10\%} & 96 / 97 \tabularnewline
Multi-Step & Base / Enh. & 1.431 / 1.396 & 82.6\% / 85.5\% & \textbf{+2.45\%} & 1.247 / 1.242 & \textbf{+0.40\%} & 91 / 92 \tabularnewline
Cyclic LR & Base / Enh. & 1.431 / 1.396 & 84.3\% / 87.4\% & \textbf{+2.46\%} & 1.268 / 1.265 & \textbf{+0.21\%} & 99 / 99 \tabularnewline
Polynomial & Base / Enh. & 1.396 / 1.381 & 83.9\% / 84.9\% & \textbf{+1.08\%} & 1.222 / \textbf{1.217} & \textbf{+0.43\%} & 96 / 98 \tabularnewline
Exponential & Base / Enh. & 1.361 / 1.345 & 88.4\% / 90.0\% & \textbf{+1.18\%} & 1.238 / 1.239 & -0.09\% & 97 / 99 \tabularnewline
Cos. Restart & Base / Enh. & 1.373 / 1.352 & 86.9\% / 89.2\% & \textbf{+1.57\%} & 1.232 / 1.236 & -0.29\% & 97 / 99 \tabularnewline
On Plateau & Base / Enh. & 1.385 / 1.359 & 85.3\% / 88.6\% & \textbf{+1.88\%} & 1.228 / 1.238 & -0.86\% & 99 / 95 \tabularnewline
\bottomrule
\end{tabular}
\end{table}

\begin{figure}[htp!]
\vspace{-40pt}
\centering
\includegraphics[width=\textwidth]{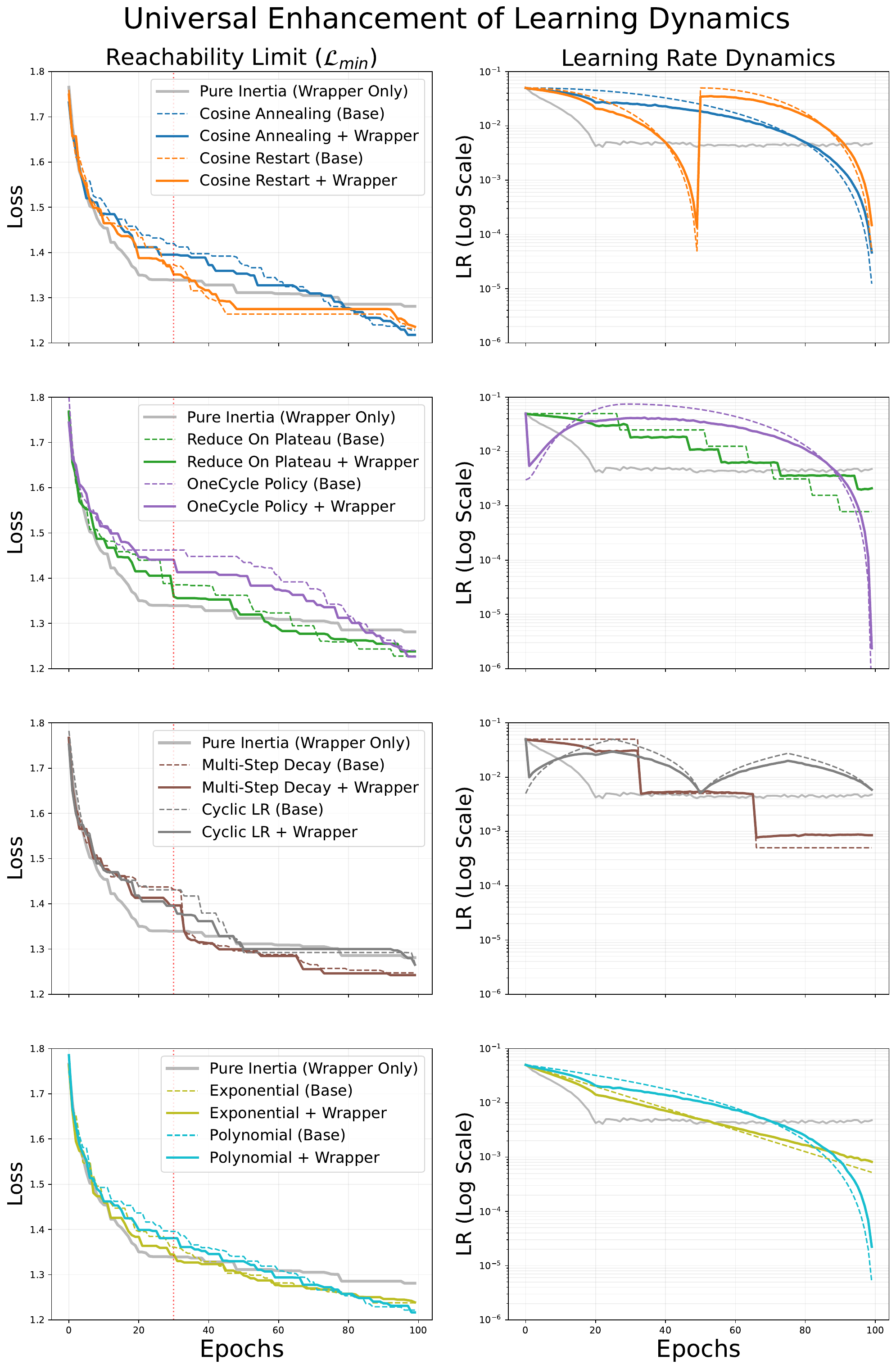}
\caption{Universal Enhancement of Learning Dynamics via the Inertia-Aware Scheduler Wrapper. The panels contrast validation loss \textbf{(Left)} and learning rate behavior \textbf{(Right)} across eight schedulers. Solid lines indicate enhanced systems, while dashed lines indicate baselines; the vertical red dotted line marks the critical Epoch 30 checkpoint where dynamical acceleration is most visible.}
\label{fig:learning_dynamics}
\end{figure}

The evolutionary trajectories of these systems are visualized in Figure~\ref{fig:learning_dynamics}, contrasting the reachability curves and learning rate behavior between the baseline and inertia-aware agents.

\paragraph{Analysis of Results}

The results from the convergence limit test reveal the profound impact of \textbf{Intelligence Inertia} on neural optimization. Below, we provide a deep attribution analysis of the observed phenomena, categorized by convergence speed, coupling logic, and the interaction between inertia-awareness and traditional heuristics.

\paragraph{1. ``Inertial Primacy'': The Rapid Convergence of Pure Inertia Group}
As shown in Table~\ref{tab:scheduler_benchmark}, the most striking performance was observed in the \textbf{Pure Inertia (Wrapper Only)} group. Despite the total absence of a pre-defined human schedule or decay timetable, the system achieved a staggering \textbf{94.31\%} convergence progress within the first 30 epochs, outperforming all native PyTorch schedulers.

\begin{itemize}
    \item \textbf{Coherence of the R-S Manifold}: This ``steepest descent'' phenomenon validates the immense power of \textbf{respecting the coherence of the R-S Manifold}. The Pure Inertia wrapper functions as a logical filter, identifying and rejecting update components that threaten the established rule-structure. In contrast, traditional schedulers maintain a high level of ``blind kinetic energy'' during early training \cite{sutskever2013importance}, consuming significant computational resources on incoherent rule-oscillations. Consequently, the convergence progress of traditional baselines lags behind the pure inertial group by approximately 10\% in the early stages.
\end{itemize}

\paragraph{2. The Rationale for a Hybrid Strategy}
While the Pure Inertia configuration demonstrated remarkable early performance, we observed the inherent limitations of a purely physical feedback system. Due to stochastic noise, the measured velocity\textbf{ ($v$)} rarely reaches an exact value of 1.0 at the final stages of convergence. This prevents the learning rate from smoothing to absolute zero as effectively as traditional policies, allowing baselines to overtake in the terminal phase.

The \textbf{Coupling Strategy} (utilizing a 20\% inertia weight) was designed to resolve this:
\begin{itemize}
    \item \textbf{Macro Inheritance}: The system inherits the ``macro-convergence strategy'' of the base scheduler, ensuring the energy level eventually reaches absolute zero.
    \item \textbf{Micro Correction}: The Scheduler wrapper performs a real-time audit on every batch. By accumulating micro-advantages through the rejection of incoherent updates while following the macro-decay trend, the system is able to push the reachability limit ($\mathcal{L}_{min}$) beyond previous theoretical expectations.
\end{itemize}

\paragraph{3. Universality of Gains and Resilience against Logic Conflicts}
The experimental data across the eight test groups confirms the robustness of the \textbf{Intelligence Inertia} framework, though it also highlights specific interactions with existing heuristics.

\begin{itemize}
    \item \textbf{Universality of Dynamical Acceleration}: In every test group, the introduction of the scheduler wrapper improved the convergence progress at Epoch 30 (indicated by the vertical red dotted line in Figure~\ref{fig:learning_dynamics}). For discrete schedulers like \textbf{Multi-Step Decay} and \textbf{Cyclic LR} (Figure~\ref{fig:learning_dynamics}, third row), the \textbf{Dyn. Accel.} exceeded 2.4\%. This indicates that the Scheduler wrapper effectively ``polishes'' the transition periods of coarse-grained algorithms, reducing energy oscillations typically caused by abrupt step changes.
    \item \textbf{Compressing the Thermodynamic Reachability Limit}: For mainstream policies like \textbf{OneCycle Policy} and \textbf{Cosine Annealing}, the wrapper successfully lowered $\mathcal{L}_{min}$. This represents \textbf{``precision etching''} at the base of the solution space—by filtering out microscopic incoherent updates, the model reaches deeper, more stable basins that are typically occluded by stochastic noise \cite{keskar2016large}.
    \item \textbf{Resilience and ``Logic Incompatibility''}: We observed slight terminal degradation in \textbf{Cosine Restart} ($-0.29\%$) and \textbf{Reduce On Plateau} ($-0.86\%$), which reveals a deep logical incompatibility between specific heuristics and physical inertia:
    \begin{itemize}
        \item \textbf{Restart Conflict} (Figure~\ref{fig:learning_dynamics}, top-right panel): WarmRestart relies on periodically injecting ``high-energy shocks'' to escape local optima. As seen in the orange curves, while the baseline allows for an unfiltered high-energy pulse, the Scheduler wrapper identifies these shocks as \textbf{Directional Dissonance} and dampens them. These two physical objectives—acceleration vs. braking—are diametrically opposed.
        \item \textbf{Plateau Paradox} (Figure~\ref{fig:learning_dynamics}, second row): \textbf{Plateau} schedulers rely on loss volatility to trigger a decay. However, the Scheduler wrapper smooths the R-S Manifold so effectively that the loss curve becomes ``too elegant,'' never reaching the threshold for a plateau. Consequently, the enhanced version maintains a higher base learning rate than the baseline, missing the opportunity for fine-grained terminal convergence.
    \end{itemize}
\end{itemize}

Crucially, these conflicts further prove the high sensitivity of the Scheduler wrapper in smoothing the information manifold. Despite this logical incompatibility, the hybrid scheme ensures that the system inherits the steepest descent properties of the loss while maintaining competitive parity with the reachability limits of the control groups.

\subsubsection{Sub-experiment II: Resilience to Noise Shocks}

To evaluate the stability of an intelligent agent in uncertain real-world environments, we subjected a 1M ResNet substrate to high-entropy informational shocks. During the latter half of the training process on CIFAR-10, we injected 100\% label noise into every other epoch, alternating between ``clean'' and ``noisy'' data streams. This experiment aims to verify whether the \textbf{Inertia-Aware Scheduler Wrapper} can maintain the purity of the system's internal rule-set through autonomous dynamical intervention, preventing the ``logical shattering'' that typically follows an entropy surge.

The comparative performance metrics for the baseline (Exponential Scheduler) and the pure inertia-aware regulation (Wrapper Only) across the shock phases are summarized in Table~\ref{tab:noise_shocks}. In this table, the following abbreviations are used: Vel. (Velocity $v$) and LR (Learning Rate).

\begin{table}[ht]
\centering
\renewcommand{\arraystretch}{1.5} 
\setlength{\tabcolsep}{5.5pt} 
\begin{tabular}{>{\centering\arraybackslash}m{0.18\textwidth}>{\centering\arraybackslash}m{0.12\textwidth}>{\centering\arraybackslash}m{0.12\textwidth}>{\centering\arraybackslash}m{0.12\textwidth}>{\centering\arraybackslash}m{0.12\textwidth}>{\centering\arraybackslash}m{0.15\textwidth}}
\toprule
\centering \textbf{Group} & \centering \textbf{Phase} & \centering \textbf{Avg. Loss} & \centering \textbf{Avg. Vel.} & \centering \textbf{Avg. LR} & \centering \textbf{Brake Ratio} \tabularnewline
\midrule
\centering Baseline (Exponential) & \centering Pre-Shock & \centering 1.4878 & \centering 0.6505 & \centering 0.041532 & \centering N/A \tabularnewline
\cmidrule{2-6}
\centering & \centering Clean Pulse & \centering 1.4142 & \centering 0.6723 & \centering 0.009068 & \centering 0.9x (None) \tabularnewline
\centering & \centering Noise Pulse & \centering 2.1845 & \centering 0.6926 & \centering 0.010367 & \centering - \tabularnewline
\midrule
\centering Pure Inertia (Wrapper Only) & \centering Pre-Shock & \centering 1.4379 & \centering 0.7525 & \centering 0.015874 & \centering N/A \tabularnewline
\cmidrule{2-6}
\centering & \centering Clean Pulse & \centering \textbf{1.3541} & \centering 0.8111 & \centering 0.001659 & \centering \textbf{1.2x} (\textbf{Active}) \tabularnewline
\centering & \centering Noise Pulse & \centering \textbf{2.0841} & \centering 0.8634 & \centering \textbf{0.001420} & \centering - \tabularnewline
\bottomrule
\end{tabular}
\caption{Comparative Dynamical Metrics under Periodic High-Entropy (100\% Noise) Shocks. This table contrasts the response of a traditional exponential scheduler against our standalone inertia-aware wrapper during intermittent noise injection, revealing that the regulated system autonomously applies protective braking while the baseline remains blind to the noise-induced entropy surge.}
\label{tab:noise_shocks}
\end{table}

The quantitative data indicates a distinct behavioral shift: while the baseline remains ``blind'' to the quality of incoming data—maintaining its learning intensity despite the lack of coherent environmental feedback—the regulated system demonstrates an immediate dynamical response. This is characterized by the autonomous modulation of the \textbf{Brake Ratio}, reflecting the system's real-time perception of structural resistance.

\paragraph{Analysis of Results}
\label{sec:logical_shock_analysis}

\begin{figure}[htbp]
    \centering
    \includegraphics[width=\textwidth]{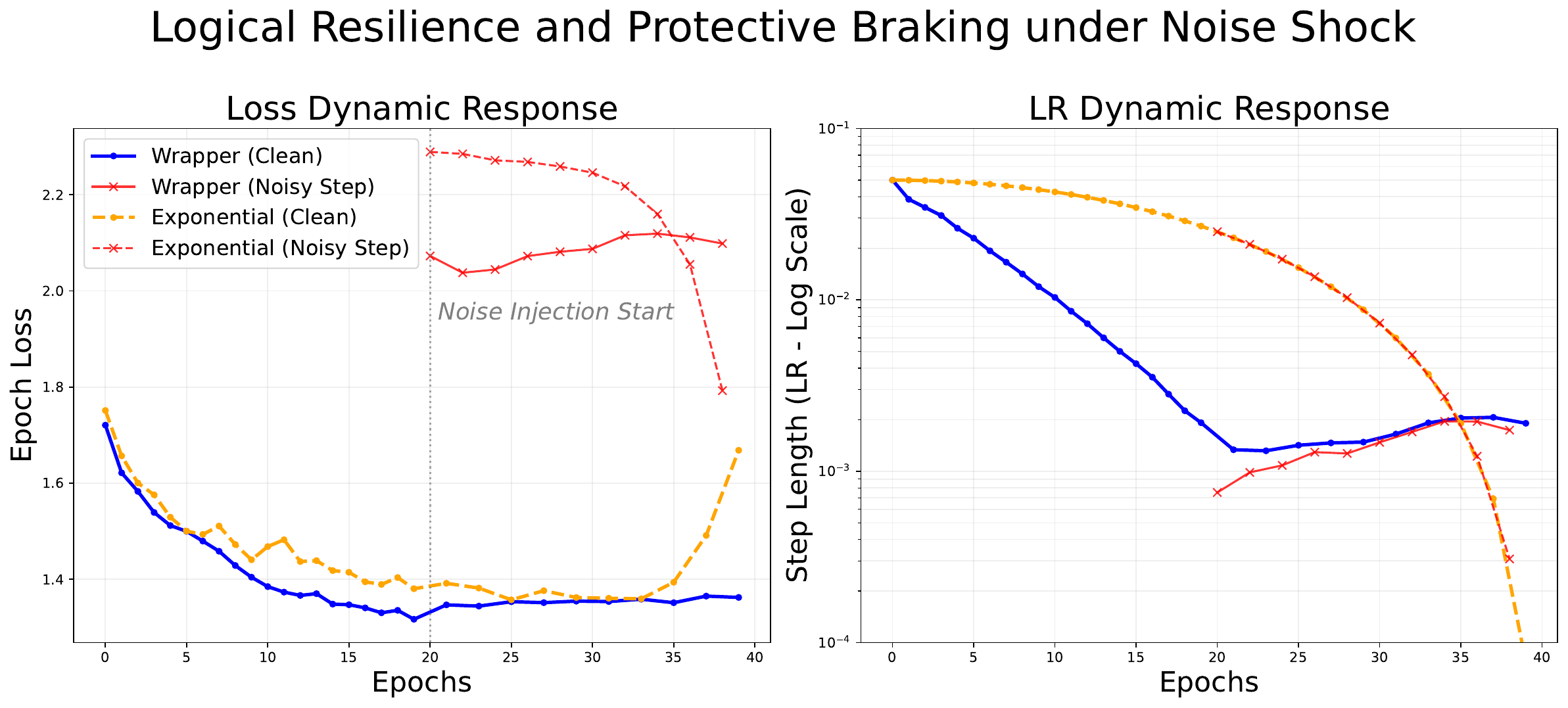}
    \caption{Logical Resilience and Relativistic Braking under Noise Shock. \textbf{(Left)} Validation loss across epochs; the baseline (dashed yellow) exhibits convergence of clean and noisy loss curves, indicating disruption of rules and assimilation by noise, while the regulated system (solid blue) maintains a distinct dual-track separation. \textbf{(Right)} Relativistic Braking response; the Scheduler wrapper autonomously suppresses its step length during noise steps to preserve rule-purity.}
    \label{fig:logic_resilience}
\end{figure}

\begin{figure}[htbp]
    \centering
    \includegraphics[width=\textwidth]{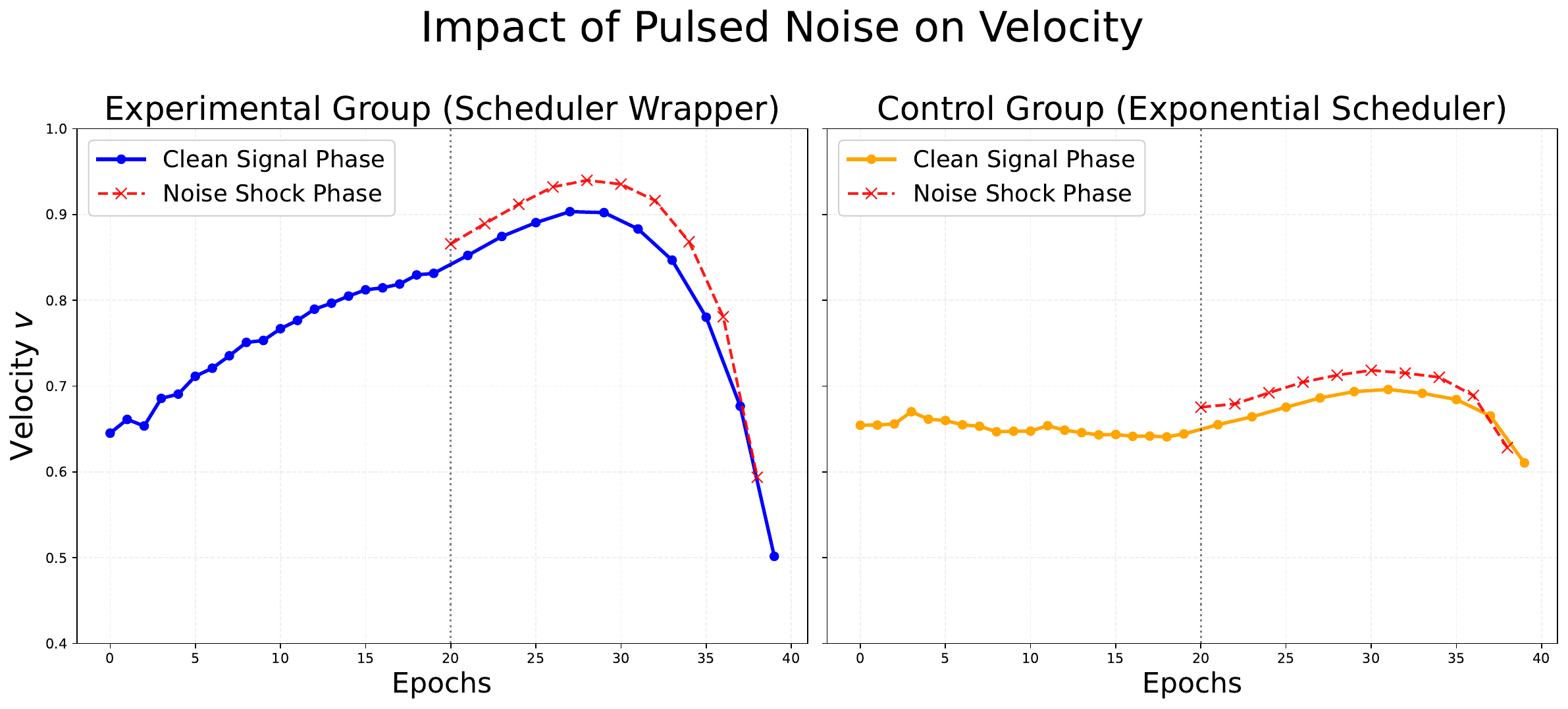}
    \caption{Impact of Pulsed Noise on Velocity. This figure illustrates the velocity response ($v$) across training epochs. \textbf{(Left)} Experimental group (Regulated) and \textbf{(Right)} Control group (Baseline). Both groups show that the onset of label noise causes a distinct phase separation, where velocity spikes from a steady-state toward the informational limit, confirming that system velocity is a reliable physical indicator of informational shocks.}
    \label{fig:velocity_separation}
\end{figure}

The results from the noise-injection experiment illustrate a fundamental distinction in how intelligent agents manage high-entropy informational surges. By integrating the quantitative metrics from Table~\ref{tab:noise_shocks} with the phenomenological trajectories in Figures~\ref{fig:logic_resilience} and~\ref{fig:velocity_separation}, we analyze the micro-mechanisms of the protective braking provided by the \textbf{Intelligence Inertia Theory}.

\paragraph{1. Quantifying the Brake Ratio and its Emergent Value}
A microscopic analysis of the indicators in Table~\ref{tab:noise_shocks} reveals that the system's resilience originates from a subtle but decisive adjustment of the brake ratio—the relative intensity of learning applied to signal versus noise. The data indicates that the regulated group’s average learning rate during noise cycles was approximately 20\% lower than during clean cycles. In an environment of 100\% noise, where infinitesimal residual alignments still exist \cite{pitelis2020label}, an absolute 100\% lock of parameters is neither optimal nor necessary.

The cumulative effect of this 1.2x braking tilt is transformative. This extra damping serves as a ``logical buffer,'' preventing noise-driven updates from penetrating the deep causal layers of the R-S Manifold. While the baseline fails because its effective strategy results in $\text{Damage} > \text{Repair}$, the regulated group achieves a stable $\text{Repair} > \text{Damage}$ state. This ``self-healing'' behavior allows models to maintain cognitive continuity even under extreme volatility.

\paragraph{2. The Physical Dynamics of Protective Braking}
Figure~\ref{fig:logic_resilience} visualizes the macroscopic consequences of this strategy on the R-S Manifold, contrasting the structural decay of a rigid scheduler with the ``self-healing'' behavior of the inertia-aware system.

In the baseline group, the loss curves for clean and noisy cycles begin to converge following the onset of noise. Because the baseline maintains nearly identical learning intensity during noise steps, the structural damage inflicted cannot be fully rectified during subsequent clean steps, leading to a cumulative buildup of entropy. Conversely, the inertia-aware group exhibits a persistent ``Dual-Track Separation,'' realizing the principle of \textbf{Controlled Damage}. By implementing an autonomous brake during high-velocity pulses, the system ensures that structural contamination remains below the recovery threshold of the subsequent repair cycle.

\paragraph{3. Velocity Manifestation of Environmental Impulses}
Figure~\ref{fig:velocity_separation} confirms that sudden velocity jumps are an endogenous physical property of intelligent agents when subjected to external shocks in the phase space.

Regardless of whether inertia-aware regulation is applied, noise injection leads to a significant and immediate leap in instantaneous velocity. This phenomenon has a direct \textbf{physical mapping}: since noise provides no logic-consistent environment feedback ($dS_{ext} \to 0$), it is functionally equivalent to a ``high-energy particle flow'' bombardment. Under this impact, the model's evolution velocity in R-S space inevitably approaches the informational limit ($v \to 1$), satisfying the conservation laws of the system's underlying dynamics. Therefore, by simply monitoring for instantaneous velocity spikes, the wrapper autonomously triggers the critical protective braking. This mechanism grants the agent \textbf{Physical Immunity}—utilizing the inherent laws of dynamics to maintain a stable ``Protect-Repair'' cycle and preserving cognitive structures in chaotic environments without manual intervention.

\subsubsection{Sub-experiment III: The Inertial Barrier in Continual Learning}
\label{sec:sub_exp_3}

To evaluate the system's resilience in one of the most demanding scenarios for an intelligent agent, we subjected a 1M-parameter ResNet substrate to a replay-free continual learning task on CIFAR-10 \cite{parisi2019continual}. This setup involves an abrupt transition from the \textbf{Old Task} (classes 0--4) to a \textbf{New Task} (classes 5--9) at Epoch 20, without utilizing any external data buffers or adjustments to the loss function. This experiment aims to verify whether an agent can utilize its intrinsic inertia ($\mu$) to generate dynamic resistance against ``logic collisions''---the sudden, high-energy conflict between the gradients of a new task and the established internal rule-set---thereby preventing the catastrophic destruction of existing knowledge.

The quantitative stability indicators captured during this transition are presented in Table~\ref{tab:continual_learning_stability}. 

\begin{table}[htb]
    \centering
    \renewcommand{\arraystretch}{1.5} 
\setlength{\tabcolsep}{5.5pt} 

    \caption{Comparison of Dynamical Stability Indicators in a Replay-Free Continual Learning Scenario. This table contrasts the baseline group against the regulated system during an abrupt task switch. The results indicate that a higher awareness of inertia leads to significantly lower forgetting and better overall task reachability by enforcing an autonomous collision brake at the moment of transition. Certain metrics are abbreviated: ``Pre-trans. Loss'' refers to Pre-transition Loss (Ep. 19) and ``Inst. Breaking Ratio'' refers to the Instantaneous Inertial Breaking Ratio.}
    \label{tab:continual_learning_stability}
    \begin{tabular}{m{0.3\textwidth}<{\centering} m{0.15\textwidth}<{\centering} m{0.13\textwidth}<{\centering} m{0.3\textwidth}<{\centering}}
        \toprule
        \textbf{Metric} & \textbf{Exponential} & \textbf{Wrapper} & \textbf{Improvement} \tabularnewline
        \midrule
        Pre-trans. Loss & 0.9253 & 0.8879 & - \tabularnewline
        Old Task Final Loss & 9.1505 & 7.8801 & \textbf{13.88\% Forgetting Reduction} \tabularnewline
        Retention Deficit ($\Delta \mathcal{L}$) & 8.2626 & 6.9922 & \textbf{15.38\% Rule Retention} \tabularnewline
        Full Task Final Loss & 4.9093 & 4.3232 & \textbf{11.94\% Comprehensive Synergy} \tabularnewline
        Inst. Breaking Ratio & 1.08x & \textbf{2.92x} & \textbf{Autonomous Braking} \tabularnewline
        \bottomrule
    \end{tabular}
\end{table}

\begin{figure}[htbp]
    \centering
    \includegraphics[width=\textwidth]{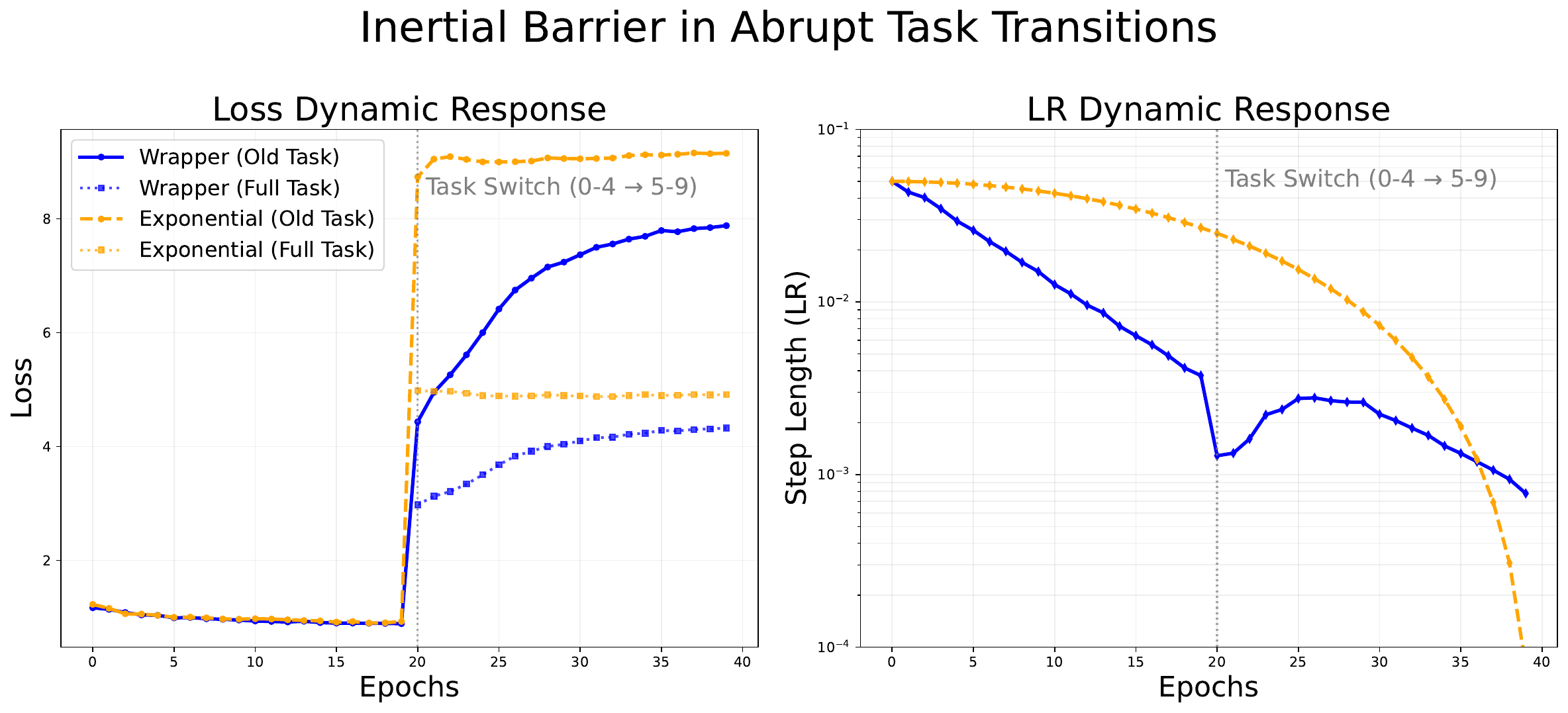}
    \caption{Inertial Barrier during Abrupt Task Transitions. This figure contrasts the behavioral strategies during the task switch at Epoch 20. \textbf{(Left)} Validation loss trajectories where the regulated group suppresses the surge of Old Task Loss; \textbf{(Right)} Step-length dynamics, where the Scheduler wrapper executes an immediate ``Protective Braking'' maneuver in response to the high-velocity logic collision.}
    \label{fig:memory_retention}
\end{figure}

The macroscopic and microscopic dynamics of this transition are visualized in Figure~\ref{fig:memory_retention}, highlighting the protective damping effect of the inertia-aware \textbf{Scheduler Wrapper}.

\paragraph{Analysis of Results}

The results of Sub-experiment III provide a decisive demonstration of how the Intelligence Inertia Theory functions as a protective barrier during abrupt task transitions. By analyzing the interplay between knowledge retention and the instantaneous dynamical response on the R-S Manifold, we identify the physical root of why inertia-aware systems survive task shifts that typically lead to the collapse of fixed-policy models.

\paragraph{1. Physical Inhibition of the Forgetting Deficit}
The \textbf{Old Task Loss} curves in Figure~\ref{fig:memory_retention} \textbf{(Left)} reveal a stark contrast in knowledge preservation between the two groups. Following the task switch at Epoch 20, the baseline group’s old task loss exhibits a near-vertical surge, peaking at $9.1505$ as shown in Table~\ref{tab:continual_learning_stability}. This reflects a catastrophic scenario where, lacking inertial resistance, the model’s established output probability distributions are physically shattered to accommodate new, non-coherent gradients. In contrast, the regulated group demonstrates remarkable \textbf{logical resilience}; the surge is significantly suppressed, ultimately locking at $7.8801$. The observed reduction in the Retention Deficit ($\Delta \mathcal{L}$) of $15.38\%$ confirms that inertia is not merely a computational burden but a fundamental shield for learned knowledge. The more an agent respects its intrinsic inertia, the more effectively it resists environmental entropy and preserves its causal integrity.

\paragraph{2. Inertia-Induced Step Contraction and Quasi-Static Assimilation}
Figure~\ref{fig:memory_retention} \textbf{(Right)} captures the instantaneous physical behavior of the learning rate at the critical transition point. The exponential baseline group fails to perceive the physical inconsistency between the existing rule-set and the new task requirements, leading to a "blind reconfiguration" that serves as the dynamical origin of catastrophic forgetting. According to the Intelligence Inertia framework, when the gradient flow of a new task deviates orthogonally from the established logic on the R-S Manifold, the rule density ($v$) approaches the informational limit. This triggers an inertia expansion, which the scheduler wrapper respects via a drastic \textbf{Relativistic Contraction} of the step length. As recorded in Table~\ref{tab:continual_learning_stability}, the regulator executed a $2.92\text{x}$ instantaneous braking, effectively providing a "mechanical buffer" that stabilizes the weight structure against non-coherent shocks. Crucially, this braking maneuver does not permanently halt learning; rather, it forces the system into a \textbf{quasi-static assimilation} mode. By taking micro-steps, the network gently weaves new task rules into the existing manifold without shattering the established topology, laying the foundation for natural unfreezing.

\paragraph{3. Robust Comprehensive Performance}
The overall performance during the task adaptation phase confirms the long-term benefits of this inertia-centric protection strategy. By the end of the experiment, the wrapper group's \textbf{Full Task Loss} was optimized by $11.94\%$ compared to the baseline. Without physical resistance, the baseline system suffers from dynamical "over-speeding," leading to irreversible logical shattering. In contrast, the regulated system ensures that the agent absorbs new knowledge in a low-dissipation manner \cite{parrondo2015thermodynamics}. Because the quasi-static assimilation protects the legacy logic, the old and new representations form a structural synergy rather than engaging in a zero-sum overwrite. This demonstrates that agents with inertial self-awareness can reach superior thermodynamic steady states, efficiently balancing the requirement for plasticity with the necessity of structural stability.

\subsubsection{Summary and Guidelines for Inertia-Aware Engineering}

Experiment III validates the engineering utility of the \textbf{Inertia-Aware Scheduler Wrapper}, transforming the theoretical framework of intelligence inertia into an autonomous \textbf{self-audit mechanism} for neural optimization. The results establish three guiding principles for the future development of resilient intelligent agents:

\begin{itemize}
    \item \textbf{Transcending Blind Search}: High optimization efficiency is achieved by identifying and suppressing ``unnecessary collisions'' between incoming gradients and the established rules on the \textbf{R-S Manifold}. By minimizing the dissipation of energy into chaotic rule-vibrations, computational work is focused exclusively on targeted structural evolution.
    \item \textbf{Objective Learning Tempo}: The learning rate is redefined as an \textbf{intrinsic physical feature} of the agent's current state rather than a pure heuristic hyperparameter. This tempo is dictated by the system's \textbf{velocity/ rule-density ($v \equiv \rho$)} and its alignment with the environmental manifold, providing a principled alternative to human-tuned decay schedules that often ignore the system's underlying \textbf{Inertia Expansion}.
    \item \textbf{Extended Applications}: Since the learning rate is but one projection of the system's \textbf{Characteristic Cycle ($l$)} onto the training process, this inertia-aware paradigm can be extended to regulate other critical variables. These include dynamic batch-size scaling based on the noise scale of the R-S Manifold, as well as automated architectural pruning and gradient clipping~\cite{mccandlish2018efficiency}.
\end{itemize}

\section{Experiment Discussion}
\label{sec:experiment_discussion}

The experiments presented in this research were designed to progressively transition our framework from theoretical hypothesis to empirical validation and engineering realization. Experiment I successfully adjudicated between classical information geometry and our relativistic formulation, confirming the presence of a ``computational wall'' and the \textbf{Inertia Expansion} effect in high-velocity informational regimes. Building upon this, Experiment II mapped the structural landscape of deep networks on the \textbf{R-S Manifold}, revealing that architectural progress is not a matter of arbitrary scaling but a constrained geometric search for an optimal, balanced trajectory. Finally, Experiment III proved the practical utility of the \textbf{Inertia-Aware Scheduler Wrapper}, demonstrating that an inertia-aware controller can stabilize and enhance learning dynamics in volatile environments by respecting the system's intrinsic physical limits.

\subsection{Theoretical Value and Potential}
The \textbf{Intelligence Inertia Theory} fills a fundamental gap in the study of machine learning by establishing a unified physical foundation for artificial intelligence dynamics:

\begin{itemize}
    \item \textbf{Bridging Discrete and Continuous Domains}: By anchoring the \textbf{Landauer limit} \cite{landauer1961} as the system's static \textbf{Rest Inertia ($\mu_0$)}, we provide a unified metric that connects foundational symbolic logic with connectionist manifold dynamics. This proves that the cost of intelligence is not a heuristic variable but is governed by thermodynamic boundaries and the \textbf{Symbolic Granularity ($\mathcal{D}$)}.
    \item \textbf{The Physics of Reachability}: Traditional optimization implicitly assumes that increasing computational power leads to proportional performance gains. Our theory exposes a hard cognitive horizon. Experiment I demonstrates that at high \textbf{velocity/rule-density ($v \equiv \rho$)}, forced reconfigurations result in a relativistic divergence of effective mass, providing the first-principles explanation for the convergence bottlenecks in network models.
    \item \textbf{A Geodesic for Structural Evolution}: Experiment II demonstrates that architectural intelligence is not a product of blind layering. Efficient evolution requires balanced coordination between \textbf{Internal Rule Reconfiguration ($dS_R$)} and \textbf{External State Gain ($dS_{ext}$)}, anchoring the trajectory near the energy equipartition point ($v \approx 0.5$). This establishes a physics-based dynamical criterion for future Neural Architecture Search (NAS) \cite{zoph2016neural}.
\end{itemize}

Furthermore, these results suggest a paradigm shift in the design of autonomous agents. By internalizing inertial feedback, future systems could evolve three profound capabilities:
\begin{itemize}
    \item \textbf{``Logic Pain''}: Intelligent agents could perceive when their velocity spikes toward the informational horizon of conflicting data, triggering spontaneous protective braking to preserve the core constants of their causal structure.
    \item \textbf{Structural Resilience}: By adopting relativistic step-length contraction, agents facing unfamiliar domains could autonomously ``freeze'' established rules, adapting to novelty through continuous, \textbf{quasi-static} integration rather than destructive reorganization \cite{parrondo2015thermodynamics}.
    \item \textbf{Energy-Optimal Evolution}: By anchoring evolution to the golden equipartition axis, agents can maximize cognitive gains with minimal entropic cost, paving the way for deeper evolution even under the constraints of noisy data and limited hardware.
\end{itemize}

\subsection{Limitations and Future Work}
While the findings demonstrate a robust unified theory, certain constraints frame the scope of this work and outline future developmental pathways:
\begin{itemize}
    \item \textbf{Scale of Validation and Large Language Models (LLMs)}: Due to the combinatorial complexity of scanning the architectural topography, our controlled empirical validations were constrained to mid-scale substrates. However, the physical principles derived herein offer a fundamental explanatory framework for the scaling behaviors observed in frontier AI. In contemporary LLMs, the massive accumulation of pre-trained knowledge pushes the core rule density toward saturation ($\rho \to 1$). Our theory predicts that forcing a global structural reconfiguration (Full Fine-Tuning) in this regime will trigger a relativistic inertia expansion, inevitably leading to catastrophic forgetting and prohibitive computational costs. 
    
    This physical prediction provides a first-principles explanation for the empirical success of Parameter-Efficient Fine-Tuning (PEFT) methods, most notably Low-Rank Adaptation (LoRA) \cite{hu2021lora}. Within the Intelligence Inertia framework, LoRA's strategy of freezing the pre-trained weights is physically equivalent to the \textbf{Relativistic Brake} ($\eta \to 0$) executed autonomously by our scheduler wrapper in Experiment III---both mechanisms serve to protect the high-inertia causal manifold from high-entropy shattering. By concurrently training a low-rank adapter, the system essentially opens a new, low-density ($v \approx 0$) parallel phase space for quasi-static assimilation. The immense stability and efficiency gains of LoRA over full fine-tuning already serve as a macroscopic empirical validation of the scalability of the inertia concept. Nevertheless, we acknowledge that designing quantitative inertial metrics specifically tailored for the attention mechanisms of Transformer architectures, and conducting formal large-scale validations, remains a critical priority for future work.

    \item \textbf{The Nature of Physical Isomorphism}: The core methodology of this paper relies on a \textbf{mathematical isomorphism} between intelligence dynamics and relativistic mechanics. It is crucial to note that this operates as an \textbf{effective theory}. Just as phonons in condensed matter physics exhibit relativistic covariant behaviors without implying the crystal lattice is the spacetime continuum \cite{volovik2003universe}, neural parameters in high-dimensional manifolds exhibit Lorentz-like cost inflation due to the structural constraints of rule-density. \textbf{We do not claim that the silicon substrate of AI agents is governed by fundamental cosmic relativity, but rather that non-commutative operator dynamics macroscopically converge to an identical algebraic form.} Future theoretical work should aim to formally derive these macroscopic emergent properties directly from microscopic stochastic gradient equations, bridging this effective physical metaphor with rigorous statistical learning theory.
    
    \item \textbf{Computational Overhead of Auditing}: Currently, the Tier-3 full-spectrum auditing protocol utilized in our experiments—executed on a single NVIDIA RTX 4070 GPU—exhibits non-negligible computational overhead. Future research will be dedicated to algorithm optimization and the development of hardware-matched libraries that compile inertial measurements into low-level operational kernels to minimize latency.
    \item \textbf{Towards Self-Referential Intelligence}: Ultimately, the Intelligence Inertia Theory posits that a true intelligent agent should not rely on external scheduling or human-tuned decay policies. Inertia-aware regulation must be deeply woven into the agent's own architecture rather than implemented as separate functions. The future of this research lies in the development of self-referential cognitive units that inherently sense and respect their own \textbf{Inertial Topology}. While realizing an agent that autonomously nurtures its own logic remains a profound challenge, the foundations laid herein provide the necessary map for this journey.
\end{itemize}

\section{Conclusion}
\label{sec:conclusion}

This paper has formalized the \textbf{physical principles} of \textit{intelligence inertia} as fundamental characteristics governing the energetic cost and dynamical stability of structural evolution in intelligent systems. By decomposing an agent into the dual operators of \textbf{Rules ($\hat{R}$)} and \textbf{States ($\hat{S}$)} and identifying their fundamental non-commutativity at the scale $\mathcal{D}$, we have moved the study of intelligence beyond phenomenological observation and into the realm of rigorous dynamical mechanics.

Our primary contribution is the derivation and empirical validation of the \textbf{Relativistic Cost Equation}. We have demonstrated, through both micro-statistical modeling of adiabatic collisions and large-scale neural experiments on the \textbf{R-S Manifold}, that the computational and entropic work required for structural reconfiguration does not scale linearly or quadratically as classical theories suggest. Instead, as an agent's trajectory approaches the limit of its symbolic interpretability, its effective mass undergoes a non-linear \textbf{Inertia Expansion}. This creates a physical ``computational wall'' that defines the absolute boundaries of reachability for any given logical substrate, effectively quantifying the physical ``weight'' of intelligence.

Crucially, by bridging the gap between abstract operator algebra and neural tensor dynamics, this work \textbf{paves a rigorous path toward the engineering realization} of inertia-aware intelligent systems. We have shown that these physical principles are highly predictive and possess immediate utility for the design of resilient agents. We have identified the \textbf{Zig-Zag Geodesic} as the optimal trajectory for architectural evolution, anchoring the system near the point of \textbf{Energy Equipartition} ($v \approx 0.5$). Furthermore, we have translated these dynamical laws into \textbf{concrete application examples}, most notably the \textbf{Inertia-Aware Scheduler Wrapper}. This practical engineering tool enables agents to achieve the steepest descent in loss while simultaneously pushing terminal reachability limits. By autonomously regulating their evolutionary tempo, models equipped with this realization exhibit a form of \textbf{``physical immunity''} against high-entropy noise and \textbf{``logical resilience''} during abrupt task transitions.

The \textbf{systematic formulation} of intelligence inertia fills a profound gap in our understanding of how complex systems learn and adapt. It provides a unified framework that bridges the thermodynamic limit of the single bit with the high-dimensional manifold dynamics of the neural circuit. Most importantly, it suggests that a true intelligent agent must not be a passive recipient of external data, but a self-aware structure that respects its own physical resistance to change. As we move toward the era of \textbf{Artificial General Intelligence (AGI)} \cite{goertzel2014artificial}, the principles established herein provide a necessary map for navigating the increasingly dense informational spacetime, paving the way for systems that are not only more powerful but fundamentally more stable, efficient, and aligned with the physical laws of the universe.

\clearpage

\section*{Declaration of Generative AI and AI-assisted technologies in the writing process}
During the preparation of this work the author(s) used \textbf{Gemini} in order to \textbf{improve language clarity, polish academic expression, and optimize the manuscript's LaTeX formatting for the journal requirements}. After using this tool/service, the author(s) reviewed and edited the content as needed and take(s) full responsibility for the content of the publication.

\clearpage

\bibliographystyle{ieeetr} 
\bibliography{references}  
\end{document}